%% file: eccv18_multimodal_arxiv.tex
\documentclass[runningheads]{llncs}
\usepackage{graphicx}
\usepackage{amsmath,amssymb} 
\usepackage{color}
\usepackage[width=122mm,left=12mm,paperwidth=146mm,height=193mm,top=12mm,paperheight=217mm]{geometry}
\usepackage[pagebackref=false,breaklinks=true,letterpaper=true,colorlinks,urlcolor=blue,citecolor=blue,linkcolor=blue,bookmarks=false]{hyperref}
\usepackage{url}
\usepackage{paralist}
\usepackage{subfig}
\usepackage{booktabs}
\usepackage{makecell}
\usepackage{gensymb}
\usepackage{pifont}
\newcommand{\cmark}{\ding{51}}%
\include{macro}

\newcommand{\final}{1}

\ifthenelse{\equal{\final}{1}}
{
\renewcommand{\jiabin}[1]{}
\renewcommand{\james}[1]{}
\renewcommand{\minghsuan}[1]{}
}
{}

\makeatletter
\newcommand{\printfnsymbol}[1]{%
  \textsuperscript{\@fnsymbol{#1}}%
}
\makeatother

\begin{document}

\pagestyle{headings}
\mainmatter
\def\ECCV18SubNumber{153}  

\title{Diverse Image-to-Image Translation via Disentangled Representations} 


\author{Hsin-Ying Lee\thanks{equal contribution}$^1$, Hung-Yu Tseng\printfnsymbol{1}$^1$, Jia-Bin Huang$^2$, Maneesh Singh$^3$, Ming-Hsuan Yang$^{1,4}$}
\institute{$^1$University of California, Merced\hspace{5pt}$^2$Virginia Tech \hspace{5pt}$^3$Verisk Analytics\hspace{5pt}$^4$Google Cloud}

\maketitle

\vspace{-7mm}
\begin{figure}[th]
\centering 
\mpage{0.535}{
\tb{Photo to van Gogh}\\
\includegraphics[width=\linewidth]{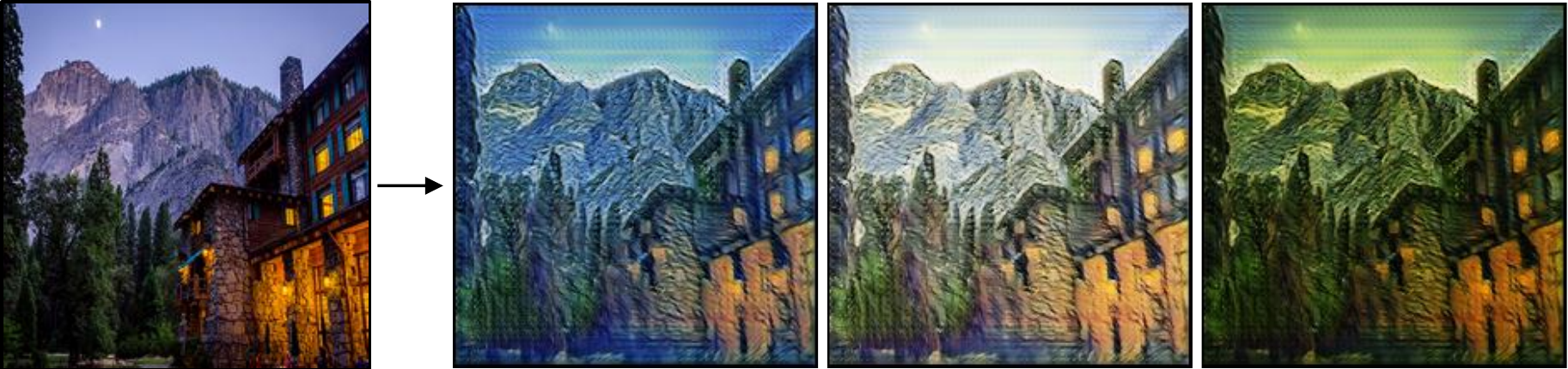} \\
\tb{Winter to summer}\\
\includegraphics[width=\linewidth]{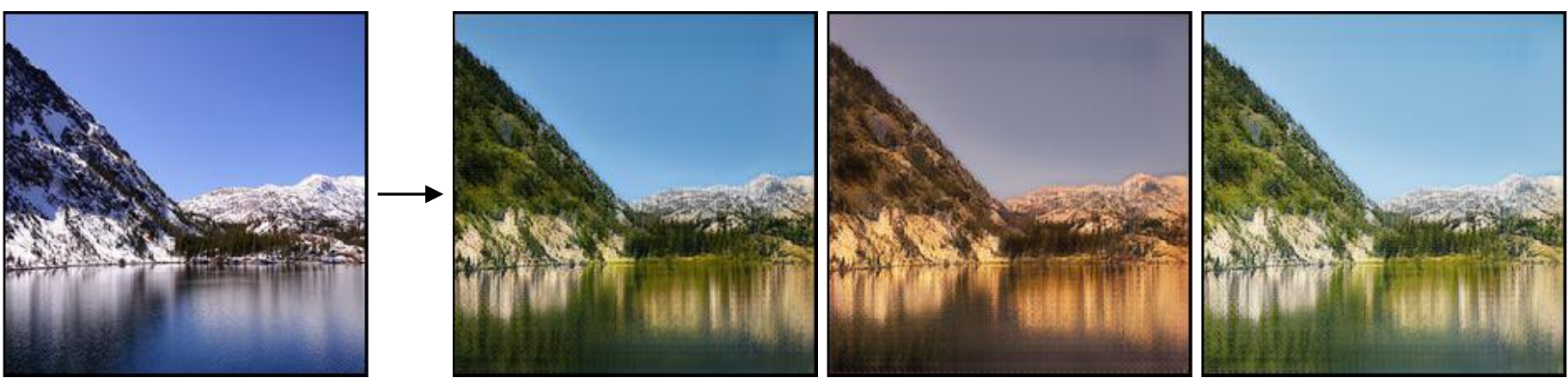} \\
\tb{Photograph to portrait}\\
\includegraphics[width=\linewidth]{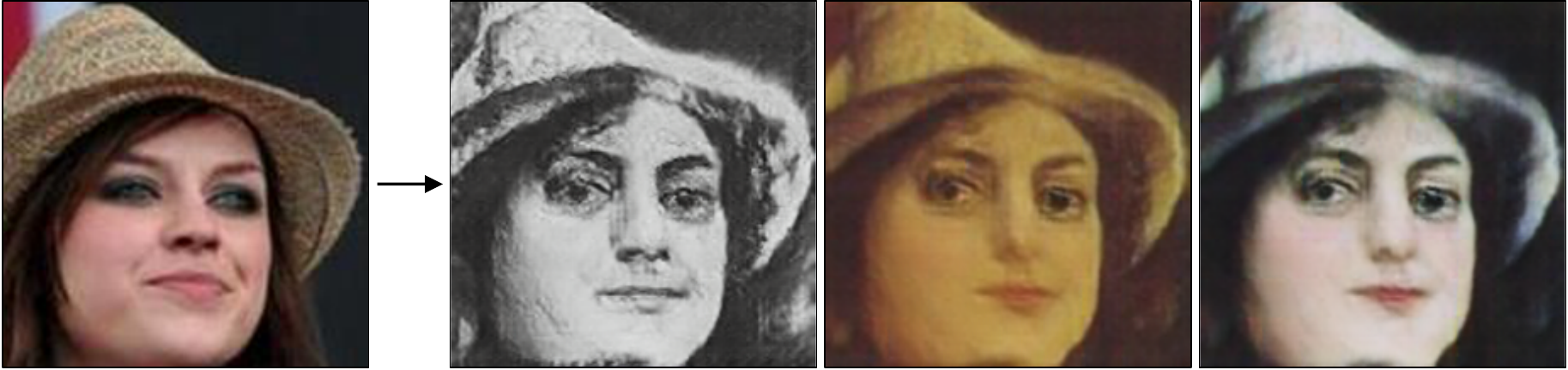} \\
 \hfill \\
} 
\hfill
\mpage{0.43}
{

\mpage{0.3}{Content}\hfill\mpage{0.3}{Attribute}\hfill\mpage{0.3}{Generated}
\includegraphics[width=\linewidth]{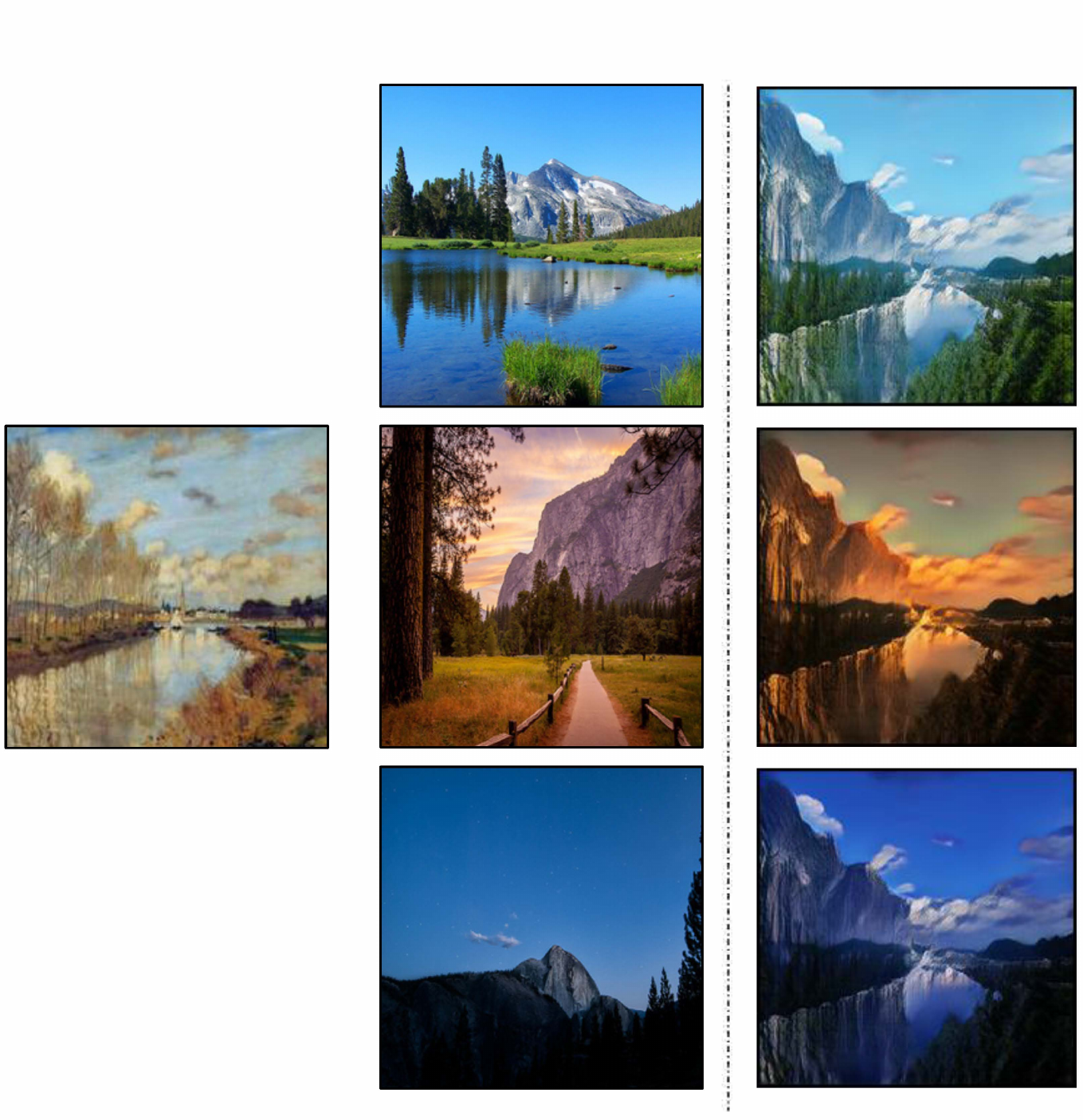} \\
}
\vspace{-5.5mm}

\mpage{0.52}{
\mpage{0.25}{Input}\hfill\mpage{0.7}{Output}
}
\hfill
\mpage{0.45}{
\mpage{0.65}{Input}\hfill\mpage{0.3}{Output}
}

 \vspace{-5mm}
    \caption{\textbf{Unpaired diverse image-to-image translation.}
    (\textit{\(Left\)}) Our model learns to perform diverse translation between two collections of images without aligned training pairs.
    (\textit{\(Right\)}) Example-guided translation.
    %
    }
    \vspace{-13mm}
    \label{figure:teaser}
\end{figure}

\begin{abstract}
Image-to-image translation aims to learn the mapping between two visual domains.
There are two main challenges for many applications: 
1) the lack of aligned training pairs and 
2) multiple possible outputs from a single input image.
In this work, we present an approach based on disentangled representation for producing diverse outputs without paired training images.
To achieve diversity, we propose to embed images onto two spaces: 
a domain-invariant content space capturing shared information across domains and a domain-specific attribute space.
Our model takes the encoded content features extracted from a given input and the attribute vectors sampled from the attribute space to produce diverse outputs at test time.
To handle unpaired training data, we introduce a novel cross-cycle consistency loss based on disentangled representations.
Qualitative results show that our model can generate diverse and realistic images on a wide range of tasks without paired training data.
For quantitative comparisons, we measure realism with user study and diversity with a perceptual distance metric.
We apply the proposed model to domain adaptation and show competitive performance when compared to the state-of-the-art on the MNIST-M and the LineMod datasets. 
%
\end{abstract}

\begin{figure}[t]
\centering
	\begin{minipage}{0.31\textwidth}
    \subfloat[CycleGAN~\cite{zhu2017cyclegan}]{
    \includegraphics[width=\linewidth]{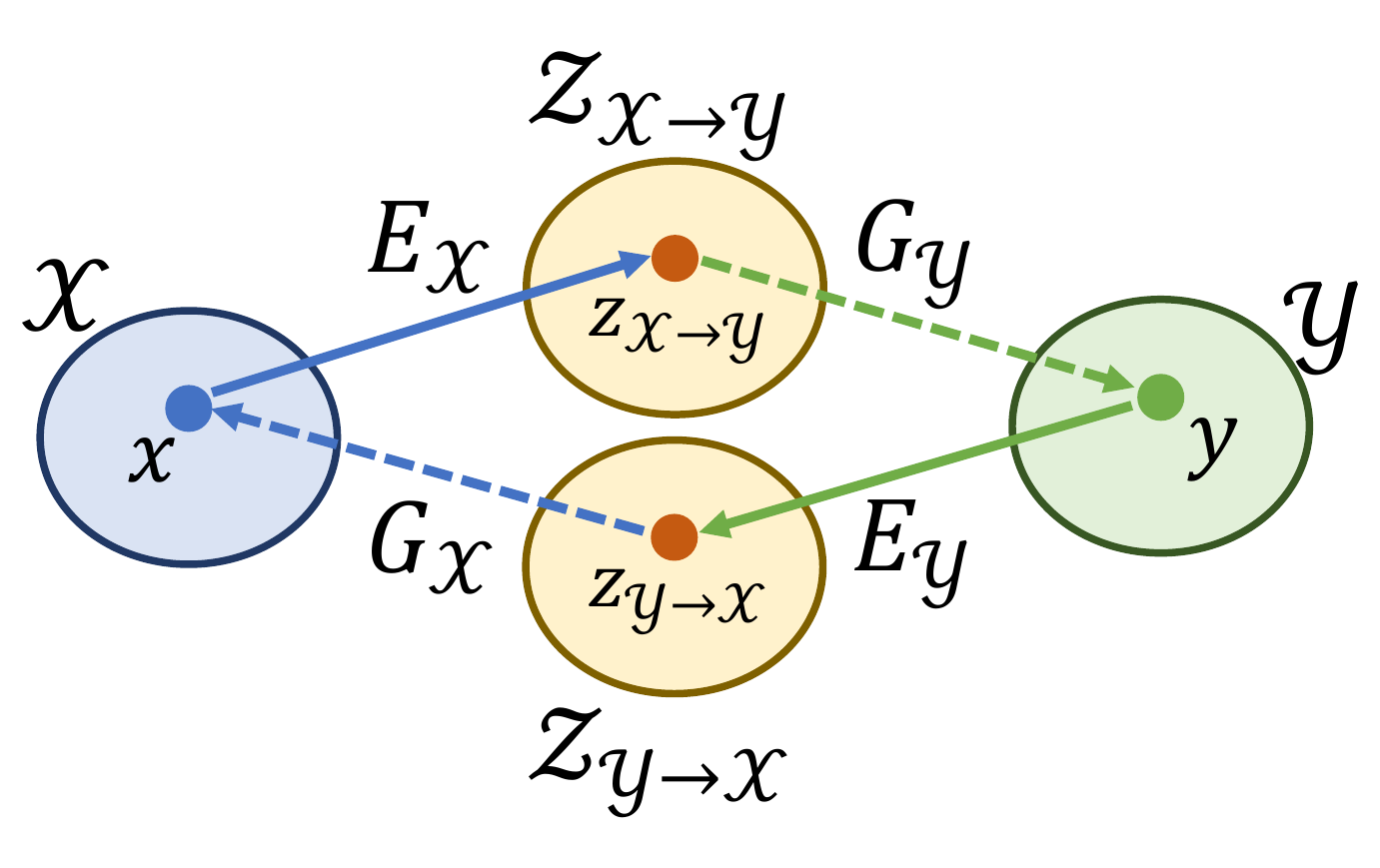}
    }
    \end{minipage}
	\hfill
    \begin{minipage}{0.31\textwidth}
	\subfloat[UNIT~\cite{liu2017unit}]{
    \includegraphics[width=\linewidth]{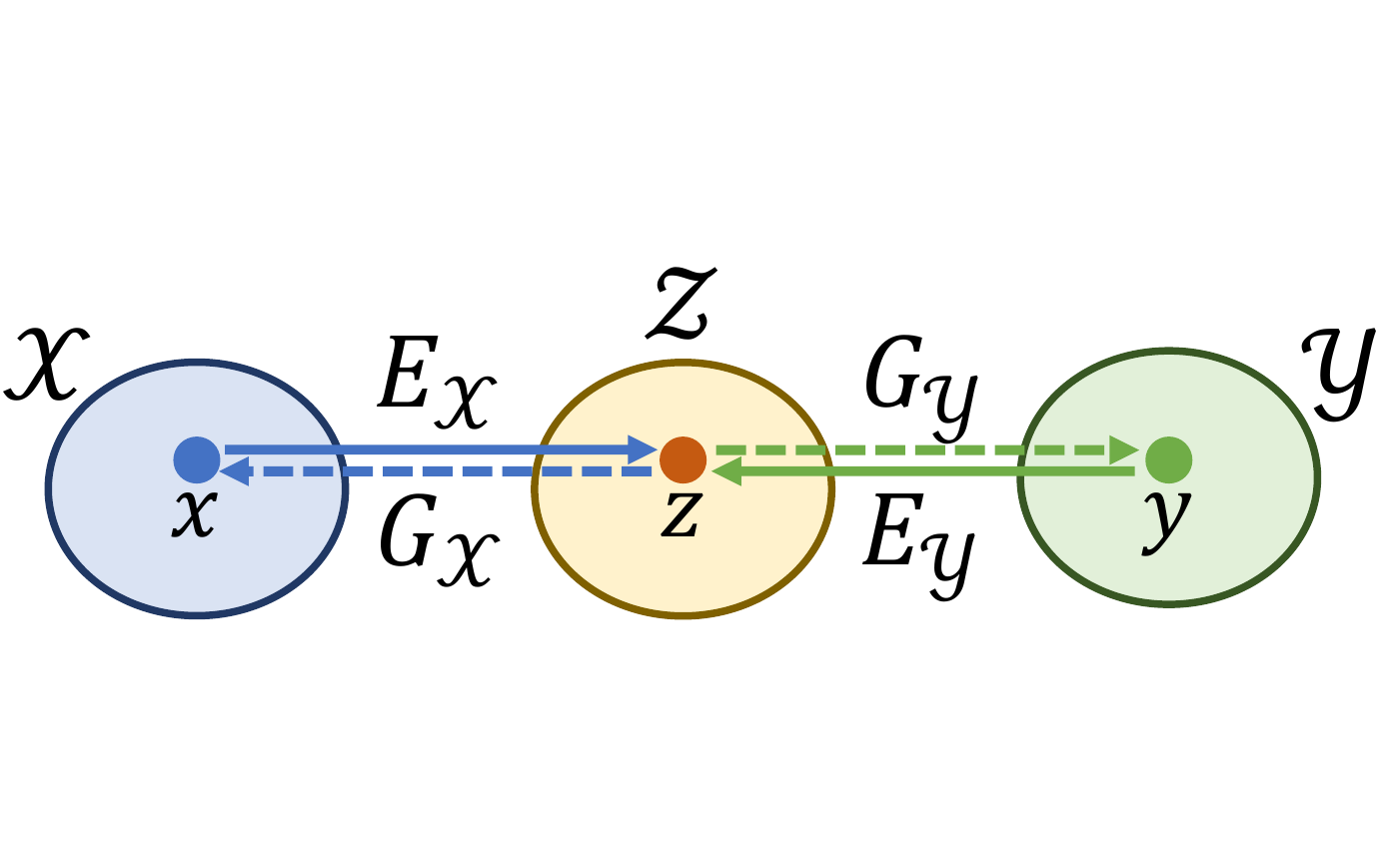}
    }
    \end{minipage}
	\hfill	
	\begin{minipage}{0.31\textwidth}
    \subfloat[Ours]{
    \includegraphics[width=\linewidth]{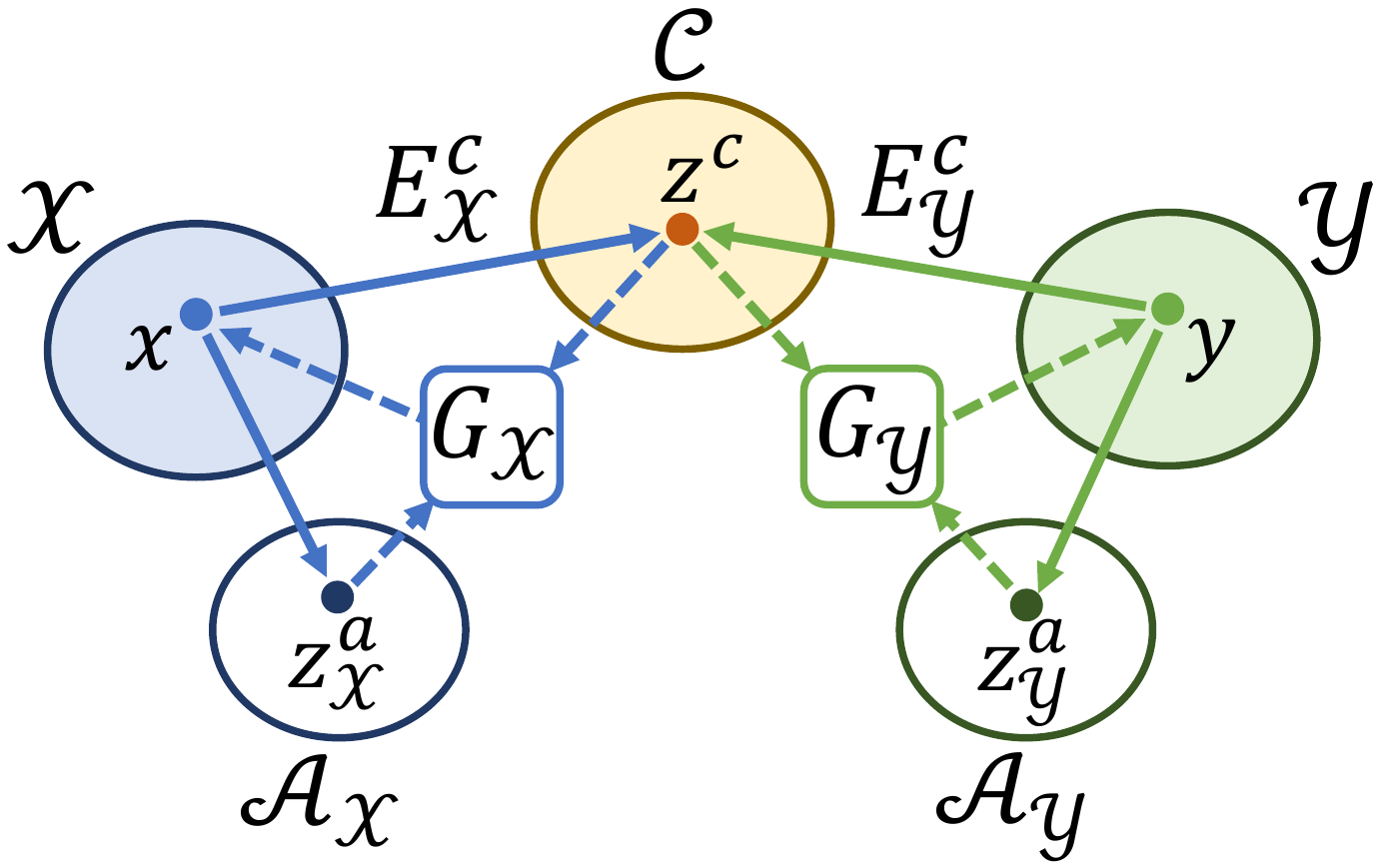}
    }
    \end{minipage}
    \caption{ 
    \textbf{Comparisons of unsupervised I2I translation methods.}
    %
    Denote $x$ and $y$ as images in domain $\mathcal{X}$ and $\mathcal{Y}$: 
    (a) CycleGAN~\cite{zhu2017cyclegan} maps $x$ and $y$ onto \emph{separated} latent spaces.
    (b) UNIT~\cite{liu2017unit} assumes $x$ and $y$ can be mapped onto a  \emph{shared} latent space.
    (c) Our approach disentangles the latent spaces of $x$ and $y$ into a shared content space $\mathcal{C}$ and an attribute space $\mathcal{A}$ of each domain.
    %
    }
    \vspace{\figmargin}
    \label{figure:assumption}
\end{figure}

\vspace{\secmargin}
\section{Introduction}
\label{sec:introduction}
\vspace{\secmargin}
Image-to-Image (I2I) translation aims to learn the mapping between different visual domains. 
Many vision and graphics problems can be formulated as I2I translation problems, such as colorization~\cite{larsson2016colorization,zhang2016colorization2} (grayscale $\rightarrow$ color), super-resolution~\cite{ledig2016photo,lai2017deep,li2016deep} (low-resolution $\rightarrow$ high-resolution), and photorealistic image synthesis~\cite{chen2017photographic,wang2017pix2pixhd} (label $\rightarrow$ image).
Furthermore, I2I translation has recently shown promising results in facilitating domain adaptation~\cite{bousmalis2017unsupervisedda,shrivastava2017apple,hoffman2017cycada,murez2018image}.

Learning the mapping between two visual domains is challenging for two main reasons. 
First, aligned training image pairs are either difficult to collect (\eg day scene $\leftrightarrow$ night scene) or do not exist (\eg artwork $\leftrightarrow$ real photo).
Second, many such mappings are inherently multimodal --- a single input may correspond to multiple possible outputs.
%
To handle multimodal translation, one possible approach is to inject a random noise vector to the generator for modeling the data distribution in the target domain. 
%
However, mode collapse may still occur easily since the generator often ignores the additional noise vectors.

Several recent efforts have been made to address these issues.
{Pix2pix}~\cite{isola2017pix2pix} applies conditional generative adversarial network to I2I translation problems.
Nevertheless, the training process requires paired data.
A number of recent work~\cite{zhu2017cyclegan,liu2017unit,yi2017dualgan,taigman2016unsupervised,choi2017stargan} relaxes the dependency on paired training data for learning I2I translation. 
%
%
These methods, however, produce a single output conditioned on the given input image.
As shown in~\cite{isola2017pix2pix,zhu2017bicyclegan}, simply incorporating noise vectors as additional inputs to the generator does not lead the increased variations of the generated outputs due to the mode collapsing issue.
The generators in these methods are inclined to overlook the added noise vectors.
Very recently, {BicycleGAN}~\cite{zhu2017bicyclegan} tackles the problem of generating diverse outputs in I2I problems by encouraging the one-to-one relationship between the output and the latent vector.
%
%
Nevertheless, the training process of {BicycleGAN} requires paired images.

In this paper, we propose a disentangled representation framework for learning to generate \emph{diverse} outputs with \emph{unpaired} training data.
Specifically, we propose to embed images onto two spaces: 
1) a domain-invariant content space and 2) a domain-specific attribute space as shown in \figref{assumption}.
%
%
%
Our generator learns to perform I2I translation conditioned on content features and a latent attribute vector. 
The domain-specific attribute space aims to model the variations within a domain given the same content, while the domain-invariant content space captures information across domains.
%
We achieve this representation disentanglement by applying a content adversarial loss to encourage the content features \emph{not} to carry domain-specific cues, and a latent regression loss to encourage the invertible mapping between the latent attribute vectors and the corresponding outputs.
%
To handle unpaired datasets, we propose a \textit{cross-cycle consistency loss} using the disentangled representations.
Given a pair of unaligned images, we first perform a cross-domain mapping to obtain intermediate results by swapping the attribute vectors from both images.
We can then reconstruct the original input image pair by applying the cross-domain mapping one more time and use the proposed cross-cycle consistency loss to enforce the consistency between the original and the reconstructed images. 
%
%
At test time, we can use either 
1) randomly sampled vectors from the attribute space to generate diverse outputs or 
2) the transferred attribute vectors extracted from existing images for example-guided translation.
%
\figref{teaser} shows examples of the two testing modes.

We evaluate the proposed model through extensive qualitative and quantitative evaluation.
%
%
In a wide variety of I2I tasks, we show diverse translation results with randomly sampled attribute vectors and example-guided translation with transferred attribute vectors from existing images.
%
%
We evaluate the realism of our results with a user study and the diversity using perceptual distance metrics~\cite{zhang2018perceptual}.
%
Furthermore, we demonstrate the potential application of unsupervised domain adaptation.
On the tasks of adapting domains from MNIST~\cite{lecun1998MNIST} to MNIST-M~\cite{ganin2016MNISTM} and Synthetic Cropped LineMod to Cropped LineMod~\cite{hinterstoisser2012linemod,wohlhart2015croplinemod}, we show competitive performance against state-of-the-art domain adaptation methods.
 

\begin{table}[t]
	\caption{\textbf{Feature-by-feature comparison of image-to-image translation networks.} Our model achieves multimodal translation without using aligned training image pairs.}
 	\label{tab:related_work}
	\centering
	\begin{tabular}{l ccccc} 
    	\toprule
Method & 
Pix2Pix~\cite{isola2017pix2pix} \ \ & CycleGAN~\cite{zhu2017cyclegan} \ \ & 
UNIT~\cite{liu2017unit} \ \ &
BicycleGAN~\cite{zhu2017bicyclegan} \ \ & 
Ours \ \  \\
        \midrule
        Unpaired& - & \cmark& \cmark&-& \cmark\\
        Multimodal&-&-&-& \cmark& \cmark\\
		\bottomrule
	\end{tabular}
    \vspace{\tabmargin}
    \vspace{-3mm}
\end{table}

We make the following contributions:

1) We introduce a disentangled representation framework for image-to-image translation. 
We apply a content discriminator to facilitate the factorization of domain-invariant content space and domain-specific attribute space, and a cross-cycle consistency loss that allows us to train the model with unpaired data.
%
%

2)
Extensive qualitative and quantitative experiments show that our model compares favorably against existing I2I models.
Images generated by our model are both diverse and realistic.

3)
We demonstrate the application of our model on unsupervised domain adaptation.
We achieve competitive results on both the MNIST-M and the Cropped LineMod datasets.

Our code, data and more results are available at \url{https://github.com/HsinYingLee/DRIT/}.

\section{Related Work}
\label{sec:related}
\vspace{-1mm}
\vspace{\secmargin}
\Paragraph{Generative adversarial networks.}
Recent years have witnessed rapid progress on generative adversarial networks (GANs)~\cite{goodfellow2014GAN,radford2016dcgan,arjovsky2017wgan} for image generation.
The core idea of GANs lies in the adversarial loss that enforces the distribution of generated images to match that of the target domain.
The generators in GANs can map from noise vectors to realistic images.
Several recent efforts explore \emph{conditional} GAN in various contexts including conditioned on text~\cite{reed2016text2img}, low-resolution images~\cite{ledig2016photo}, video frames~\cite{vondrick2016videogan}, and image~\cite{isola2017pix2pix}.
%
Our work focuses on using GAN conditioned on an input image.
In contrast to several existing conditional GAN frameworks that require paired training data, our model produces diverse outputs without paired data.
This suggests that our method has wider applicability to problems where paired training datasets are scarce or not available.

\vspace{\paramargin}
\Paragraph{Image-to-image translation.}
I2I translation aims to learn the mapping from a source image domain to a target image domain.
%
%
Pix2pix~\cite{isola2017pix2pix} applies a conditional GAN to model the mapping function.
Although high-quality results have been shown, the model training requires paired training data. 
To train with unpaired data, CycleGAN~\cite{zhu2017cyclegan}, DiscoGAN~\cite{kim2017discogan}, and UNIT~\cite{liu2017unit} leverage cycle consistency to regularize the training.
However, these methods perform generation conditioned solely on an input image and thus produce one single output.
Simply injecting a noise vector to a generator is usually not an effective solution to achieve multimodal generation due to the lack of regularization between the noise vectors and the target domain. 
%
On the other hand, BicycleGAN~\cite{zhu2017bicyclegan} enforces the bijection mapping between the latent and target space to tackle the mode collapse problem.
Nevertheless, the method is only applicable to problems with paired training data. 
Table~\ref{tab:related_work} shows a feature-by-feature comparison among various I2I models. 
Unlike existing work, our method enables I2I translation with diverse outputs in the absence of paired training data.

Very recently, several concurrent works~\cite{almahairi2018augmented,huang2018munit,cao2018dida,ma2018exemplar} (all independently developed) also adopt a disentangled representation similar to our work for learning diverse I2I translation from unpaired training data.
We encourage the readers to review these works for a complete picture.
%
%
\vspace{\paramargin}

\Paragraph{Disentangled representations.}
The task of learning disentangled representation aims at modeling the factors of data variations.
Previous work makes use of labeled data to factorize representations into class-related and class-independent components~\cite{cheung2014discovering,kingma2014semi,makhzani2015adversarial,mathieu2016disentangling}.
Recently, the unsupervised setting has been explored~\cite{chen2016infogan,denton2017unsupervised}.
InfoGAN~\cite{chen2016infogan} achieves disentanglement by maximizing the mutual information between latent variables and data variation.
Similar to DrNet~\cite{denton2017unsupervised} that separates time-independent and time-varying components with an adversarial loss, we apply a content adversarial loss to disentangle an image into domain-invariant and domain-specific representations to facilitate learning diverse cross-domain mappings. 

\vspace{\paramargin}
\Paragraph{Domain adaptation.}
Domain adaptation techniques focus on addressing the domain-shift problem between a source and a target domain.
Domain Adversarial Neural Network (DANN)~\cite{ganin2015unsupervised,ganin2016domain} and its variants~\cite{tzeng2014deep,bousmalis2016domain,Tsai_adaptseg_2018} tackle domain adaptation through learning domain-invariant features.
%
Sun~\etal~\cite{sun2016return} aims to map features in the source domain to those in the target domain. 
I2I translation has been recently applied to produce simulated images in the target domain by translating images from the source domain~\cite{ganin2015unsupervised,hoffman2017cycada}.
%
Different from the aforementioned I2I based domain adaptation algorithms, our method does not utilize source domain annotations for I2I translation. 
%

\begin{figure*}[t]
	\centering
	\subfloat[Training with unpaired images]{%
		\includegraphics[width=0.95\linewidth]{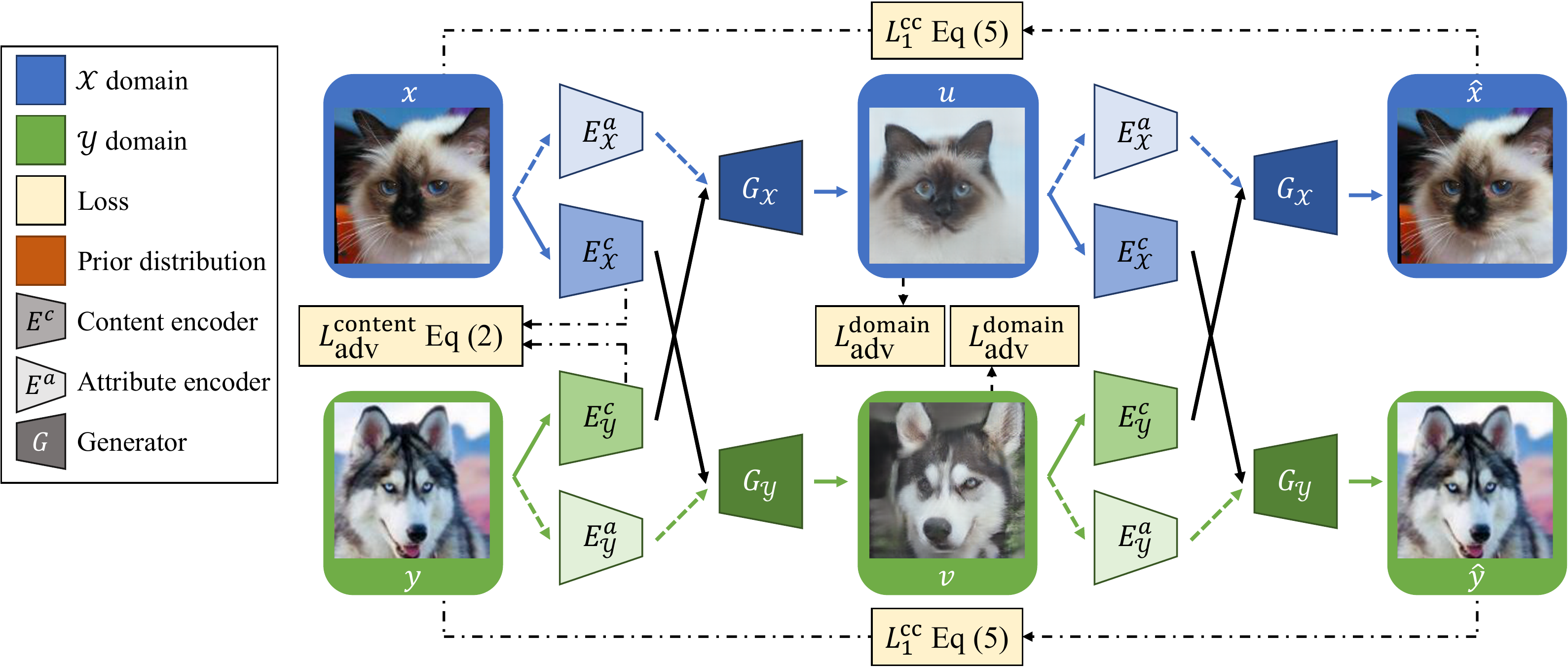}%
	}
    
    \begin{minipage}[b]{0.491\textwidth}
    \subfloat[Testing with random attributes]{%
		\includegraphics[width=\linewidth]{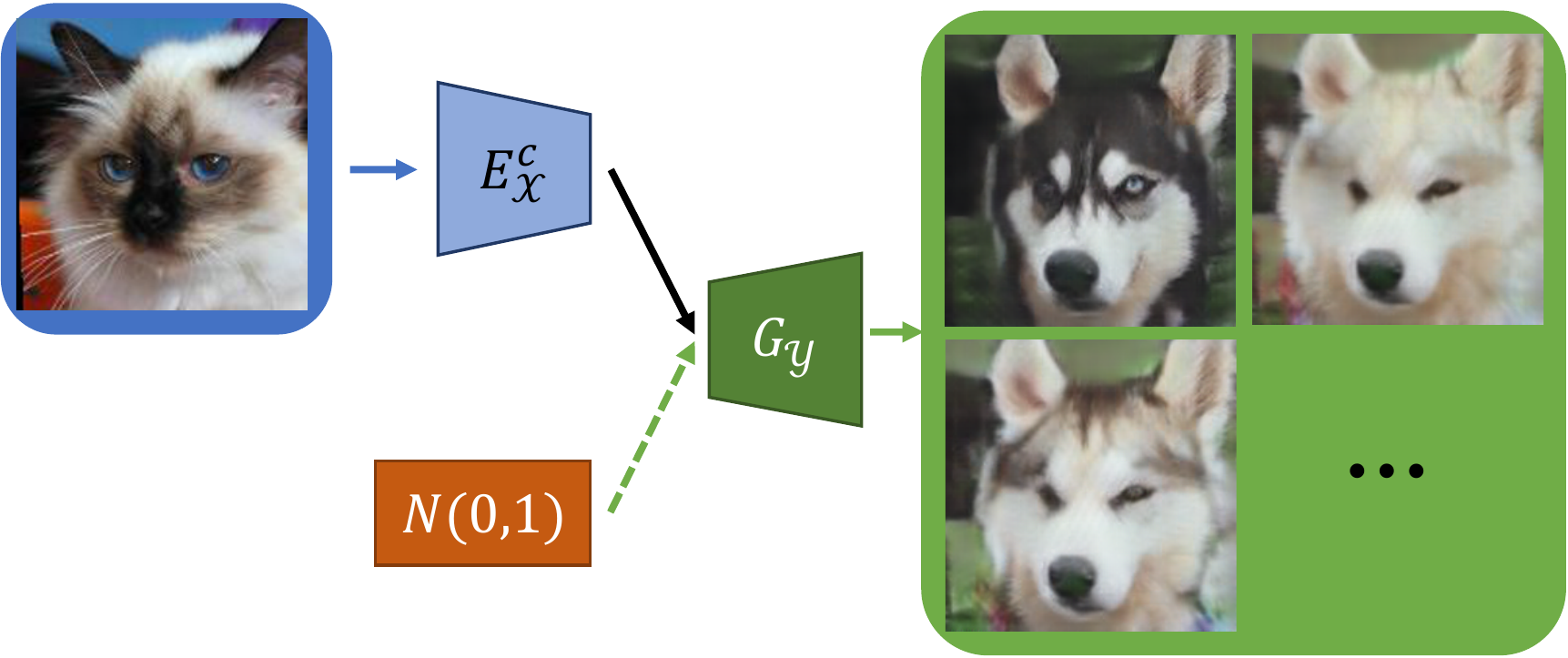}%
	}
    \end{minipage}
    \hspace{4mm}
    \begin{minipage}[b]{0.409\textwidth}
    \subfloat[Testing with a given attribute ]{%
		\includegraphics[width=\linewidth]{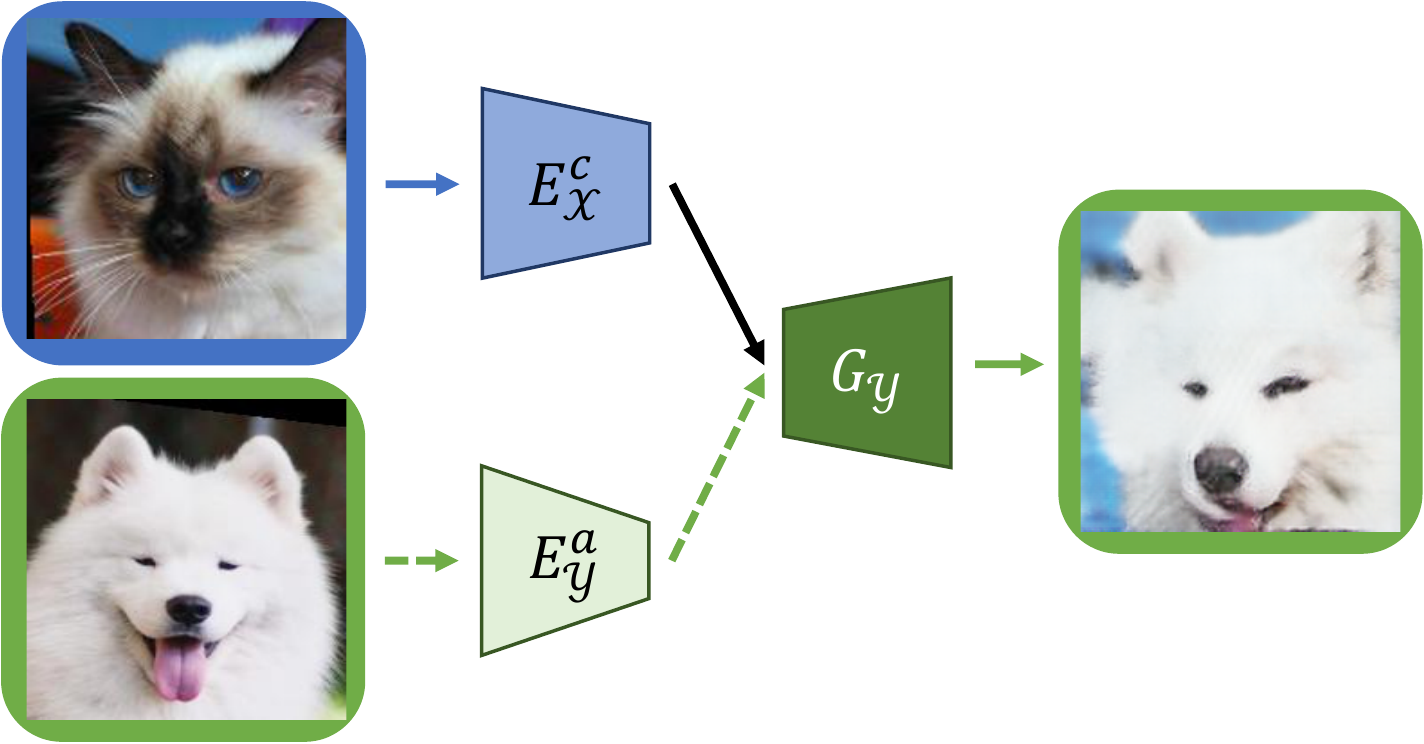}%
	}
    \end{minipage}
    \caption{\textbf{Method overview.} (a) With the proposed content adversarial loss $L_\mathrm{adv}^\mathrm{content}$ (Section~\ref{subsec:conadv}) and the cross-cycle consistency loss $L_1^\mathrm{cc}$ (Section~\ref{subsec:crosscycle}), we are able to learn the multimodal mapping between the domain $\mathcal{X}$ and $\mathcal{Y}$ with unpaired data. 
    Thanks to the proposed disentangled representation, we can generate output images conditioned on either (b) random attributes or (c) a given attribute at test time.}
    \label{figure:architecture}
    \vspace{\figmargin}
\end{figure*}


\section{Disentangled Representation for I2I Translation}
\label{sec:framework}
\vspace{\secmargin}
Our goal is to learn a multimodal mapping between two visual domains $\mathcal{X} \subset \mathbb{R}^{H\times W \times 3}$ and $\mathcal{Y} \subset \mathbb{R}^{H\times W \times 3}$ without paired training data.
As illustrated in~\figref{architecture}, our framework consists of content encoders $\{E^c_\mathcal{X}, E^c_\mathcal{Y}\}$, attribute encoders $\{E^a_\mathcal{X}, E^a_\mathcal{Y}\}$, generators $\{G_\mathcal{X}, G_\mathcal{Y}\}$, and domain discriminators $\{D_\mathcal{X}, D_\mathcal{Y}\}$ for both domains, and a content discriminators $D_\mathrm{adv}^\mathrm{c}$.
Take domain $\mathcal{X}$ as an example, the content encoder $E^c_\mathcal{X}$ maps images onto a shared, domain-invariant content space ($E^c_\mathcal{X}:\mathcal{X}\to \mathcal{C}$) and the attribute encoder $E^a_\mathcal{X}$ maps images onto a domain-specific attribute space ($E^a_\mathcal{X}:\mathcal{X}\to \mathcal{A}_\mathcal{X}$).
The generator $G_\mathcal{X}$ generates images conditioned on both content and attribute vectors ($G_\mathcal{X}:\{\mathcal{C}, \mathcal{A}_\mathcal{X}\} \to \mathcal{X} $).
The discriminator $D_\mathcal{X}$ aims to discriminate between real images and translated images in the domain $\mathcal{X}$.
Content discriminator $D^c$ is trained to distinguish the extracted content representations between two domains.
To enable multimodal generation at test time, we regularize the attribute vectors so that they can be drawn from a prior Gaussian distribution $\mathnormal{N}(0,1)$. 

In this section, we first discuss the strategies used to disentangle the content and attribute representations in \subsecref{conadv} and then introduce the proposed cross-cycle consistency loss that enables the training on unpaired data in \subsecref{crosscycle}. 
Finally, we detail the loss functions in \subsecref{learn}.

\subsection{Disentangle Content and Attribute Representations}
\label{subsec:conadv}
\vspace{\subsecmargin}
Our approach embeds input images onto a shared content space $\mathcal{C}$, and domain-specific attribute spaces, $\mathcal{A}_\mathcal{X}$ and $\mathcal{A}_\mathcal{Y}$.
Intuitively, the content encoders should encode the common information that is \emph{shared} between domains onto $\mathcal{C}$, while the attribute encoders should map the remaining domain-specific information onto $\mathcal{A}_\mathcal{X}$ and $\mathcal{A}_\mathcal{Y}$.
\vspace{\eqmargin}
\begin{equation}
\begin{aligned}
&\{z_x^{c},z_x^{a}\} = \{{E^c_\mathcal{X}}(x), {E^a_\mathcal{X}}(x)\}\quad&& z_x^{c}\in \mathcal{C}, z_x^{a}\in \mathcal{A}_\mathcal{X}\\
&\{z_y^{c},z_y^{a}\} = \{{E^c_\mathcal{Y}}(y), {E^a_\mathcal{Y}}(y)\}\quad&& z_y^{c}\in \mathcal{C}, z_y^{a}\in \mathcal{A}_\mathcal{Y}
\end{aligned}
\end{equation}
\vspace{\eqmargin}

To achieve representation disentanglement, we apply two strategies: weight-sharing and a content discriminator.
First, similar to~\cite{liu2017unit}, based on the assumption that two domains share a common latent space, we share the weight between the last layer of $E^c_\mathcal{X}$ and $E^c_\mathcal{Y}$ and the first layer of $G_\mathcal{X}$ and $G_\mathcal{Y}$.
Through weight sharing, we force the content representation to be mapped onto the same space.
However, sharing the same high-level mapping functions cannot guarantee the same content representations encode the same information for both domains.
%
Therefore, we propose a content discriminator $D^c$ which aims to distinguish the domain membership of the encoded content features $z_x^{c}$ and $z_y^{c}$.
On the other hand, content encoders learn to produce encoded content representations whose domain membership cannot be distinguished by the content discriminator $D^c$.
We express this content adversarial loss as:
\vspace{\eqmargin}
\begin{equation}
\begin{aligned}
L_{\mathrm{adv}}^{\mathrm{content}}(E^c_\mathcal{X},E^c_\mathcal{Y}, D^c) &= \mathbb{E}_{x}[\frac{1}{2}\log{D^c(E^c_\mathcal{X}(x))}+\frac{1}{2}\log{(1-D^c(E^c_\mathcal{X}(x)))]}\\  &+ \mathbb{E}_{y}[\frac{1}{2}\log{D^c(E^c_\mathcal{Y}(y))}+\frac{1}{2}\log{(1-D^c(E^c_\mathcal{Y}(y)))}]
\end{aligned}
\end{equation}
\vspace{\eqmargin}
\vspace{\eqmargin}

\vspace{\subsecmargin}
\subsection{Cross-cycle Consistency Loss}
\label{subsec:crosscycle}
\vspace{\subsecmargin}
%
With the disentangled representation where the content space is shared among domains and the attribute space encodes intra-domain variations, we can perform I2I translation by combining a content representation from an arbitrary image and an attribute representation from an image of the target domain.
We leverage this property and propose a \textit{cross-cycle consistency}.
In contrast to cycle consistency constraint in~\cite{zhu2017cyclegan} (\ie{$\mathcal{X} \to \mathcal{Y} \to \mathcal{X}$}) which assumes one-to-one mapping between the two domains, the proposed cross-cycle constraint exploit the disentangled content and attribute representations for cyclic reconstruction.
%

Our cross-cycle constraint consists of two stages of I2I translation.

\noindent \textbf{Forward translation.} Given a non-corresponding pair of images $x$ and $y$,  we encode them into $\{z_x^{c}, z_x^{a}\}$ and $\{z_y^{c}, z_y^{a}\}$.
We then perform the first translation by swapping the attribute representation (\ie $z_x^{a}$ and $z_y^{a}$) to generate $\{u,v\}$, where $u\in \mathcal{X}, v \in \mathcal{Y}$.
\vspace{\eqmargin}
\begin{equation}
\begin{aligned}
u = G_\mathcal{X}(z_y^c, z_x^a)\quad
v = G_\mathcal{Y}(z_x^c, z_y^a)
\end{aligned}
\end{equation}
\vspace{\eqmargin}

\noindent \textbf{Backward translation.} After encoding $u$ and $v$ into $\{z_u^c,z_u^a\}$ and $\{z_v^c,z_v^a\}$, we perform the second translation by once again swapping the attribute representation (\ie $z_u^a$ and $z_v^a$).
\vspace{\eqmargin}
\begin{equation}
\begin{aligned}[c]
\hat{x} = G_\mathcal{X}(z_v^c, z_u^a)\quad
\hat{y} = G_\mathcal{Y}(z_u^c, z_v^a)
\end{aligned}
\end{equation}
\vspace{\eqmargin}

Here, after two I2I translation stages, the translation should reconstruct the original images $x$ and $y$ (as illustrated in \figref{architecture}).
To enforce this constraint, we formulate the \textit{cross-cycle consistency loss} as:

\vspace{\eqmargin}
\begin{equation}
\begin{aligned}
L_1^{\mathrm{cc}}(G_\mathcal{X},G_\mathcal{Y},E_\mathcal{X}^c,E_\mathcal{Y}^c,E_\mathcal{X}^a,E_\mathcal{Y}^a) = 
\mathbb{E}_{x,y}[&\lVert G_\mathcal{X}(E_\mathcal{Y}^c(v),E_\mathcal{X}^a(u) )-x \lVert_{1} \\
+ &\lVert G_\mathcal{Y}(E_\mathcal{X}^c(u),E_\mathcal{Y}^a(v) )-y \lVert_{1}],\\
\end{aligned}
\end{equation}
\vspace{\eqmargin}
where $u=G_\mathcal{X}(E_\mathcal{Y}^c(y)),E_\mathcal{X}^a(x))$ and $v=G_\mathcal{Y}(E_\mathcal{X}^c(x)),E_\mathcal{Y}^a(y))$.

\begin{figure*}[t]
	\centering
		\includegraphics[width=\linewidth]{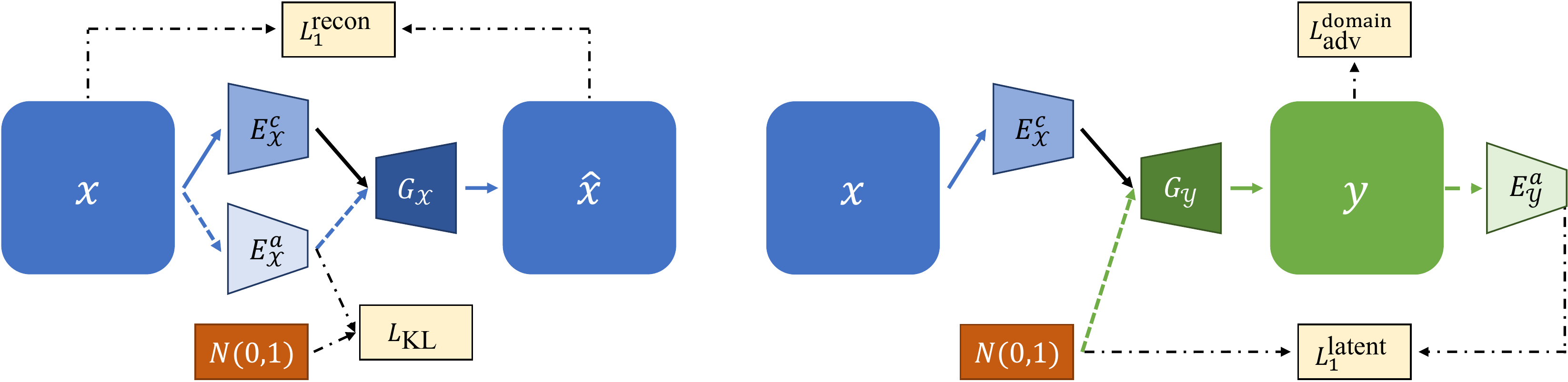}%
	\caption{\textbf{Loss functions.} 
    In addition to the cross-cycle reconstruction loss $L_1^{\mathrm{cc}}$ and the content adversarial loss $L_{\mathrm{adv}}^\mathrm{content}$ described in Figure~\ref{figure:architecture}, we apply several additional loss functions in our training process. 
    The self-reconstruction loss $L_1^{\mathrm{recon}}$ facilitates training with self-reconstruction;
    the KL loss $L_{\mathrm{KL}}$ aims to align the attribute representation with a prior Gaussian distribution;
    the adversarial loss $L_{\mathrm{adv}}^{\mathrm{domain}}$ encourages $G$ to generate realistic images in each domain; 
    and the latent regression loss $L_1^{\mathrm{latent}}$ enforces the reconstruction on the latent attribute vector.
    More details can be found in Section~\ref{subsec:learn}.}
	\label{figure:loss}
    \vspace{\figmargin}
\end{figure*}

\subsection{Other Loss Functions}
\label{subsec:learn}
\vspace{\subsecmargin}
Other than the proposed content adversarial loss and cross-cycle consistency loss, we also use several other loss functions to facilitate network training.
We illustrate these additional losses in \figref{loss}. 
Starting from the top-right, in the counter-clockwise order:

\noindent \textbf{Domain adversarial loss.}
We impose adversarial loss $L_{\mathrm{adv}}^{\mathrm{domain}}$ where $D_\mathcal{X}$ and $D_\mathcal{Y}$ attempt to discriminate between real images and generated images in each domain, while $G_\mathcal{X}$ and $G_\mathcal{Y}$ attempt to generate realistic images.
%

\noindent \textbf{Self-reconstruction loss.}
In addition to the cross-cycle reconstruction, we apply a self-reconstruction loss $L_1^{\mathrm{rec}}$ to facilitate the training. 
With encoded content/attribute features $\{z_x^c, z_x^a\}$ and $\{z_y^c, z_y^a\}$, the decoders $G_\mathcal{X}$ and $G_\mathcal{Y}$ should decode them back to original input $x$ and $y$.
That is, $\hat{x} = G_\mathcal{X}(E_\mathcal{X}^c(x),E_\mathcal{X}^a(x) )$ and $\hat{y} = G_\mathcal{Y}(E_\mathcal{Y}^c(y),E_\mathcal{Y}^a(y) )$.
%

\noindent \textbf{KL loss.}
In order to perform stochastic sampling at test time, we encourage the attribute representation to be as close to a prior Gaussian distribution.  
We thus apply the loss $L_{\mathrm{KL}}= \mathbb{E}[D_{\mathrm{KL}}((z_a)\|N(0,1))]$, where $D_{\mathrm{KL}}(p\|q)=-\int{p(z)\log{\frac{p(z)}{q(z)}}\mathrm{d}z}$.

\noindent \textbf{Latent regression loss.}
To encourage invertible mapping between the image and the latent space, we apply a latent regression loss $L_1^{\mathrm{latent}}$ similar to~\cite{zhu2017bicyclegan}. 
We draw a latent vector $z$ from the prior Gaussian distribution as the attribute representation and attempt to reconstruct it with $\hat{z}=E_\mathcal{X}^a(G_\mathcal{X}(E_\mathcal{X}^c(x),z))$ and $\hat{z}=E_\mathcal{Y}^a(G_\mathcal{Y}(E_\mathcal{Y}^c(y),z))$.
%
%
%
%

\begin{figure*}[t]
	\centering
     \mpage{0.16}{Input}\hfill\mpage{0.8}{Generated images}
    \includegraphics[width=\linewidth]{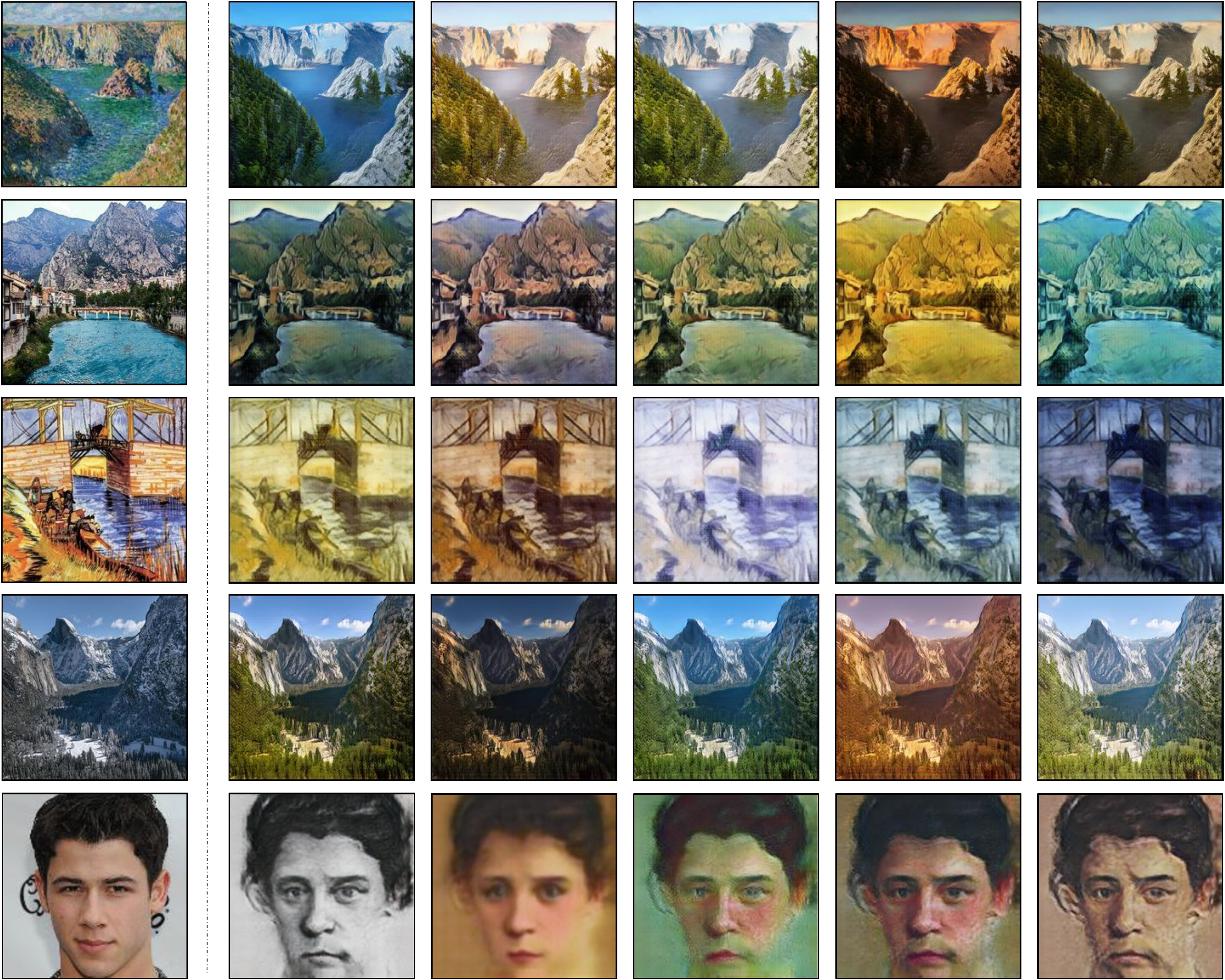}
    \caption{\textbf{Sample results.} We show example results produced by our model. The left column shows the input images in the source domain. The other five columns show the output images generated by sampling random vectors in the attribute space. 
    The mappings from top to bottom are: Monet $\rightarrow$ photo, photo $\rightarrow$ van Gogh, van Gogh  $\rightarrow$ Monet, winter $\rightarrow$ summer, and photograph $\rightarrow$ portrait.
    }
    \label{figure:example}
    \vspace{\figmargin}
 \end{figure*}

The full objective function of our network is:
\vspace{\eqmargin}
\begin{equation}
\begin{aligned}
\min_{G,E^c,E^a}\max_{D,D^c}\quad &\lambda_{\mathrm{adv}}^{\mathrm{content}}L_{\mathrm{adv}}^{\mathrm{c}}+\lambda_1^{\mathrm{cc}}L_1^{\mathrm{cc}} + \lambda_{\mathrm{adv}}^{\mathrm{domain}}L_{\mathrm{adv}}^{\mathrm{domain}}+ \lambda_1^{\mathrm{recon}} L_1^{\mathrm{recon}}\\
 + &\lambda_1^{\mathrm{latent}}L_1^{\mathrm{latent}}+ \lambda_{\mathrm{KL}}L_{\mathrm{KL}}
\end{aligned}
\end{equation}
\vspace{\eqmargin}
where the hyper-parameters $\lambda$s control the importance of each term. 
%

%

\begin{figure*}[t]
    \subfloat{%
	    \includegraphics[width=\linewidth]{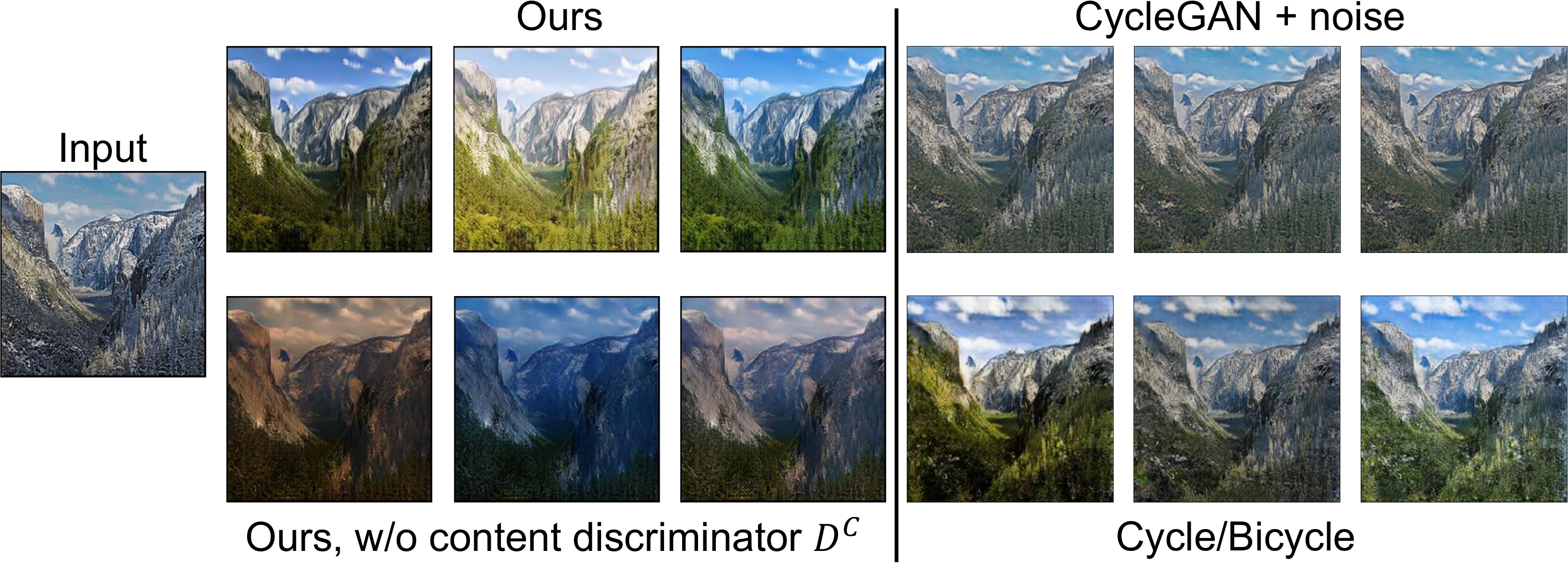}%
	}
	\caption{\textbf{Diversity comparison.} On the winter $\rightarrow$ summer translation task, our model produces more diverse and realistic samples over baselines.
	}
	\label{figure:diversity}
    \vspace{-4mm}
\end{figure*}
\begin{figure*}[t]
	\centering
	\subfloat{%
	    \includegraphics[width=\linewidth, height=3cm]{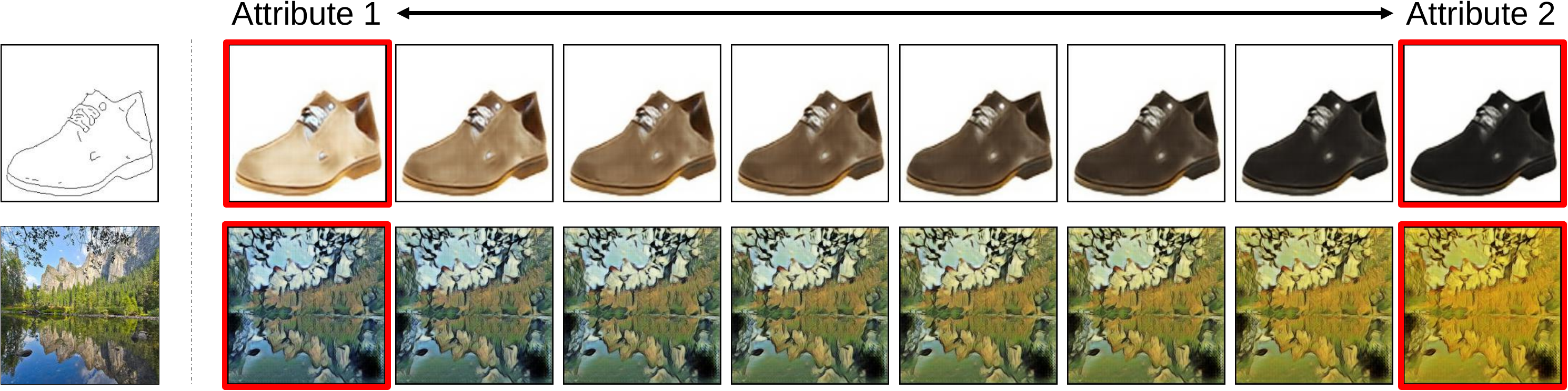}%
	}
	\caption{\textbf{Linear interpolation between two attribute vectors.}  Translation results with linear-interpolated attribute vectors between two attributes (highlighted in red).
	}
	\label{figure:interpolation}
    \vspace{\figmargin}
\end{figure*}
\vspace{\secmargin}
\section{Experimental Results}
\label{sec:experiments}
\vspace{\secmargin}
\vspace{-1mm}
\Paragraph{Implementation details.}
We implement our model with PyTorch~\cite{paszke2017pytorch}.
We use the input image size of $216 \times 216$ for all of our experiments except domain adaptation.
For the content encoder $E^c$, we use an architecture consisting of three convolution layers followed by four residual blocks.
For the attribute encoder $E^a$, we use a CNN architecture with four convolution layers followed by fully-connected layers.
We set the size of the attribute vector to $z^a \in R^8$ for all experiments.
For the generator $G$, we use an architecture containing four residual blocks followed by three fractionally strided convolution layers.
For more details of architecture design, please refer to the supplementary material.

For training, we use the Adam optimizer~\cite{kinga2015adam} with a batch size of $1$, a learning rate of $0.0001$, and exponential decay rates $(\beta_1, \beta_2) = (0.5, 0.999)$.
In all experiments, we set the hyper-parameters as follows: $\lambda^{\mathrm{content}}_{\mathrm{adv}}=1$, $ \lambda_{\mathrm{cc}}=10$, $\lambda^{\mathrm{domain}}_{\mathrm{adv}}=1$, $ \lambda_1^{\mathrm{rec}} =10$,  $\lambda_1^{\mathrm{latent}}=10$, and $\lambda_{\mathrm{KL}}=0.01$. 
We also apply an L1 weight regularization on the content representation with a weight of $0.01$. 
We follow the procedure in DCGAN~\cite{radford2016dcgan} for training the model with adversarial loss.

\vspace{\paramargin}
\Paragraph{Datasets.}
We evaluate our model on several datasets include Yosemite~\cite{zhu2017cyclegan} (summer and winter scenes), artworks~\cite{zhu2017cyclegan} (Monet and van Gogh), edge-to-shoes~\cite{Yu2014edge2shoe} and photo-to-portrait cropped from subsets of the WikiArt dataset \footnote{\url{https://www.wikiart.org/}} and the CelebA dataset~\cite{liu2015celeb}. 
We also perform domain adaptation on the classification task with MNIST~\cite{lecun1998MNIST} to MNIST-M~\cite{ganin2016MNISTM}, and on the classification and pose estimation tasks with Synthetic Cropped LineMod to Cropped LineMod~\cite{hinterstoisser2012linemod,wohlhart2015croplinemod}.

\vspace{-3mm}
\vspace{\paramargin}
\Paragraph{Compared methods.}
We perform the evaluation on the following algorithms:
\begin{compactitem}
\item \tb{DRIT: } We refer to our proposed model, Disentangled Representation for Image-to-Image Translation, as DRIT.
\item \tb{DRIT w/o $D^c$: } Our proposed model without the content discriminator.
\item\tb{CycleGAN}~\cite{zhu2017cyclegan}, \tb{UNIT}~\cite{liu2017unit}, \tb{BicycleGAN}~\cite{zhu2017bicyclegan}
\item \tb{Cycle/Bicycle: } As there is no previous work addressing the problem of multimodal generation from unpaired training data, we construct a baseline using a combination of CylceGAN and BicycleGAN. Here, we first train CycleGAN on unpaired data to generate corresponding images as \emph{pseudo} image pairs. We then use this pseudo paired data to train BicycleGAN.
\end{compactitem}

\begin{figure}[t]
	\centering
	\subfloat[Inter-domain attribute transfer]{%
    	\mpage{0.45}{
    		\mpage{0.28}{Content}\hfill\mpage{0.28}{Attribute}\hfill\mpage{0.3}{Output}
			\includegraphics[width=\linewidth]{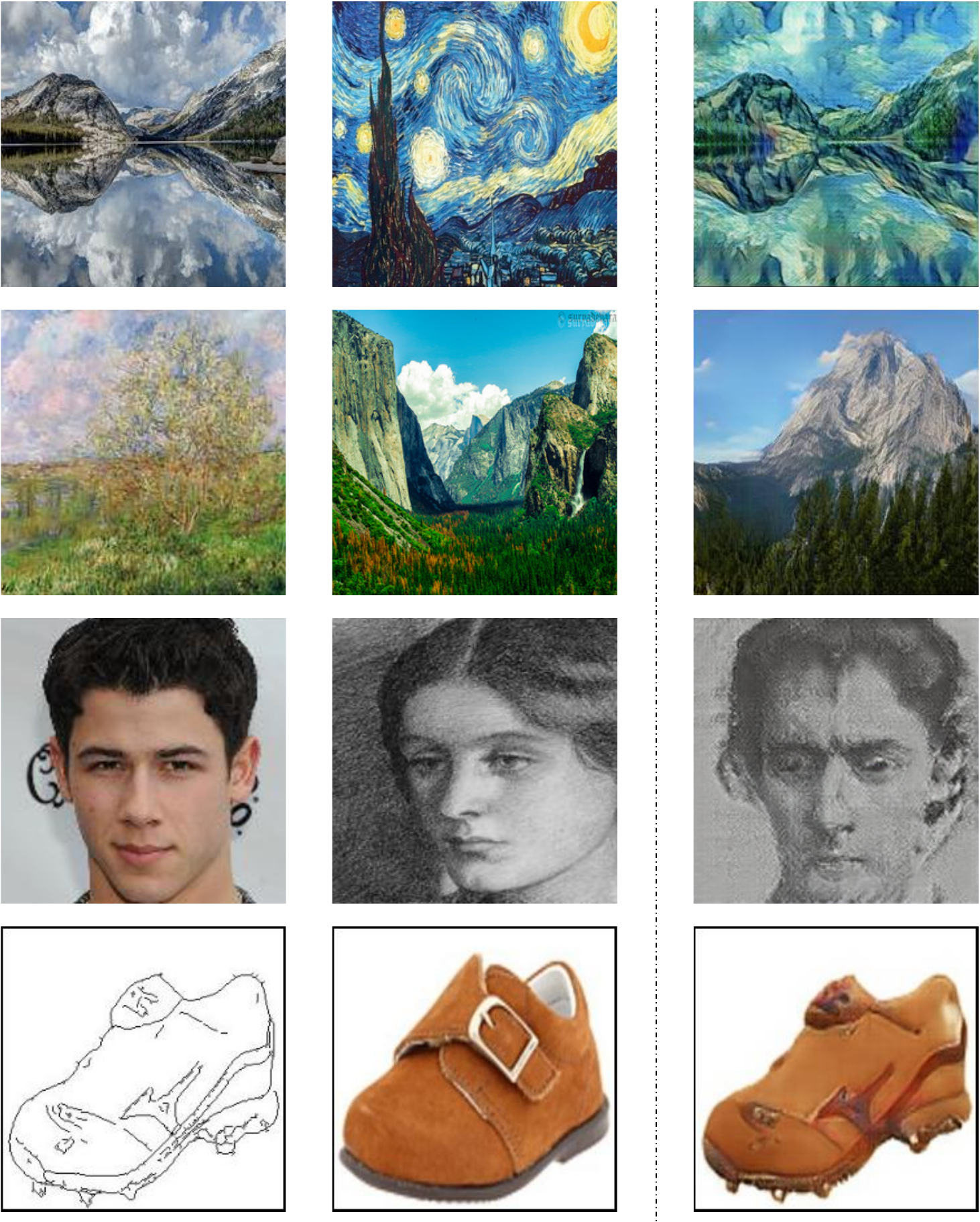}%
        }
     }
    \hfill
    \subfloat[Intra-domain attribute transfer]{%
          \mpage{0.45}{
     	\mpage{0.28}{Content}\hfill\mpage{0.28}{Attribute}\hfill\mpage{0.3}{Output}
		\includegraphics[width=\linewidth]{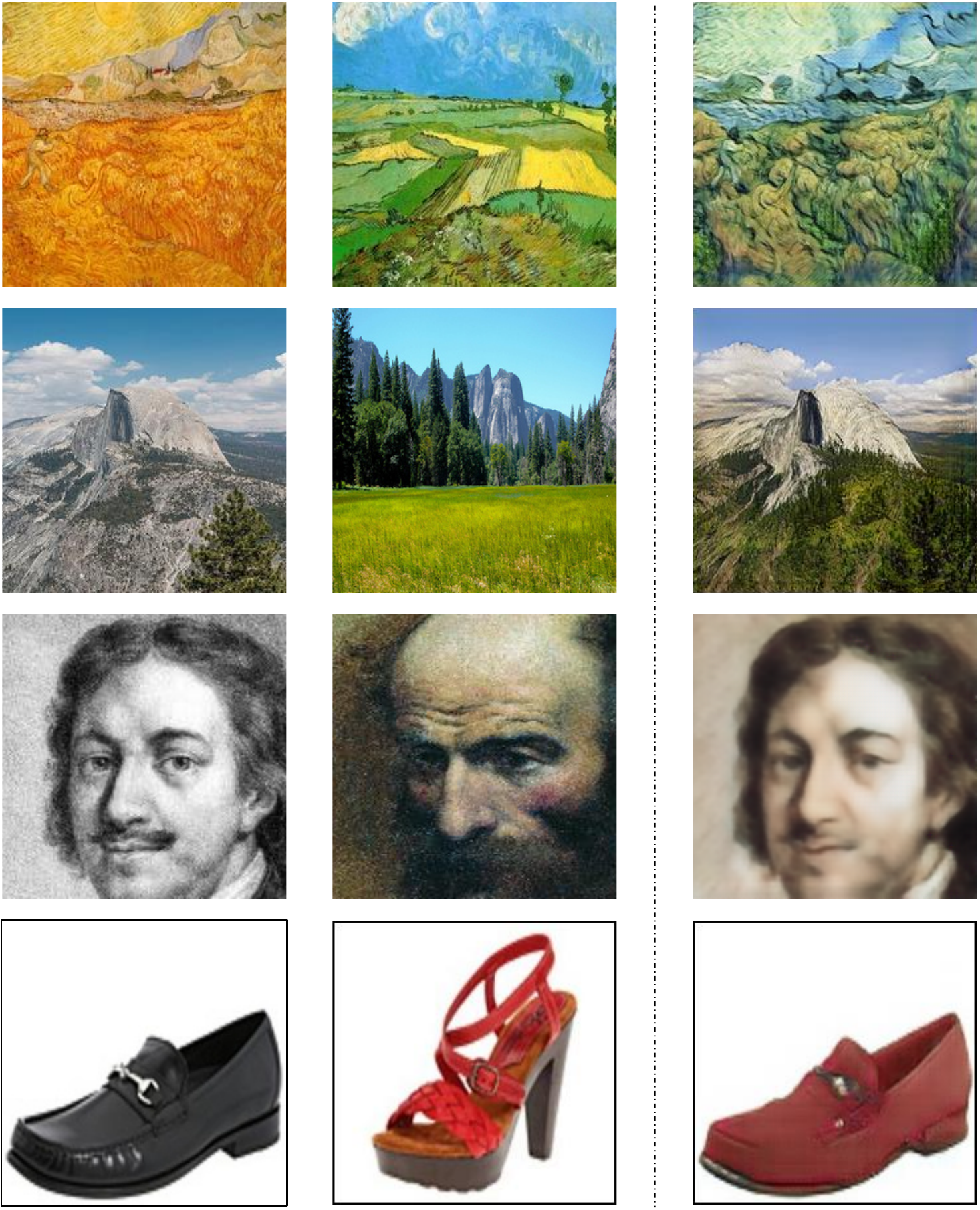}%
	   }
    }
	\caption{\textbf{Attribute transfer.} At test time, in addition to random sampling from the attribute space, we can also perform translation with the query images with the desired attributes. Since the content space is shared across the two domains, we not only can achieve (a) inter-domain, but also (b) intra-domain attribute transfer. Note that we do not explicitly involve intra-domain attribute transfer during training.
	}
	\label{figure:cross}
    \vspace{\figmargin}
\end{figure}

\begin{figure*}[t]
	\centering
    \includegraphics[width=\linewidth]{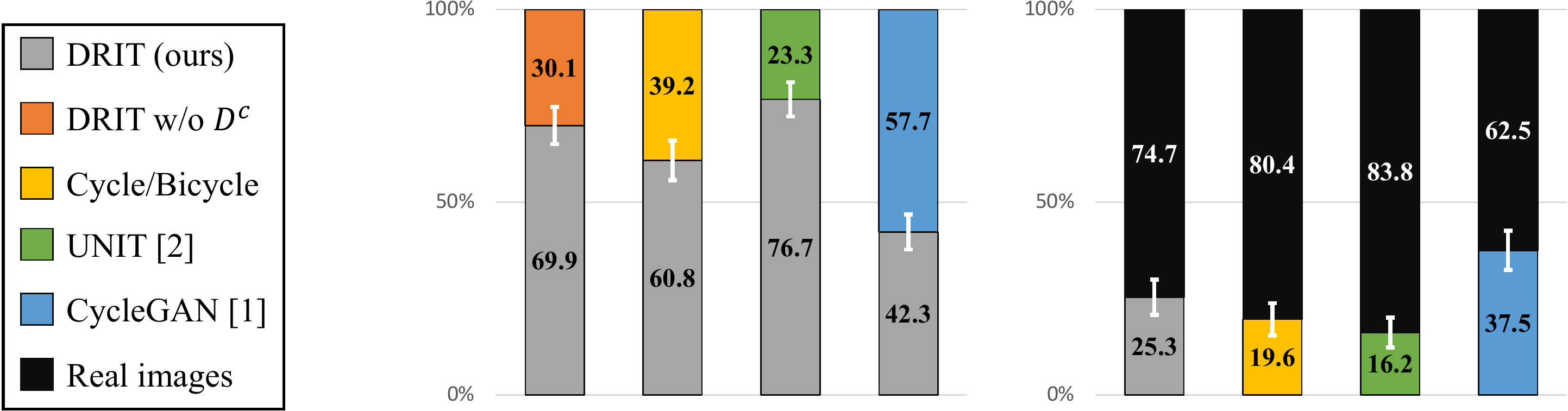}
    \caption{\textbf{Realism preference results.} 
    We conduct a user study to ask subjects to select results that are \emph{more realistic} through pairwise comparisons.
    The number indicates the percentage of preference on that comparison pair. 
    We use the winter $\rightarrow$ summer translation on the Yosemite dataset for this experiment.}
    \label{figure:realism}
    \vspace{-5mm}
 \end{figure*}
\begin{table}[tb]
    \begin{minipage}[t]{.45\linewidth}
      \caption{\textbf{Diversity.} We use the LPIPS metric~\cite{zhang2018perceptual} to measure the diversity of generated images on the Yosemite dataset. 
      }
	\label{tab:diversity}
      \centering
    \begin{tabular}{l c}
    \toprule
    Method & Diversity \\
    \midrule
    real images & .448 $\pm$ .012 \\
    \midrule
    DRIT &  \textbf{.424} $\pm$ .010 \\
    DRIT w/o $D^c$ & .410  $\pm$ .016\\
    UNIT~\cite{liu2017unit} & .406 $\pm$ .022\\
    CycleGAN~\cite{zhu2017cyclegan} & \underline{.413} $\pm$ .008\\
    Cycle/Bicycle & .399 $\pm$ .009\\
    \bottomrule
    \end{tabular}
    \end{minipage}%
    \hfill
    \begin{minipage}[t]{.45\linewidth}
        \caption{\textbf{Reconstruct error}. We use the edge-to-shoes dataset to measure the quality of our attribute encoding. The reconstruction error is $\lVert y-G_\mathcal{Y}(E_\mathcal{X}^c(x), E_\mathcal{Y}^a(y))\lVert_{1}$. * BicycleGAN uses \emph{paired} data for training.
}
	\label{tab:recon}
    \centering
	\begin{tabular}{l cc} 
    	\toprule
		Method & Reconstruct error\\
        \midrule
        BicycleGAN~\cite{zhu2017bicyclegan}* & \textbf{0.0945} \\
        \midrule 
        DRIT & \underline{0.1347}\\
        DRIT,  w/o $D^c$& 0.2076\\
		\bottomrule
	\end{tabular}
    \end{minipage} 
    \vspace{-1mm}
\end{table}

\subsection{Qualitative Evaluation}
\vspace{-1mm}
\Paragraph{Diversity.} We first demonstrate the diversity of the generated images on several different tasks in \figref{example}.
In \figref{diversity}, we compare the proposed model with other methods. 
Both our model without $D^c$ and Cycle/Bicycle can generate diverse results. 
However, the results contain clearly visible artifacts. 
Without the content discriminator, our model fails to capture domain-related details (\eg the color of tree and sky).
Therefore, the variations take place in global color difference.
%
Cycle/Bicycle is trained on pseudo paired data generated by CycleGAN.
The quality of the pseudo paired data is not uniformly ideal.
As a result, the generated images are of ill-quality.
%
%
%
%

To have a better understanding of the learned domain-specific attribute space, we perform linear interpolation between two given attributes and generate the corresponding images as shown in \figref{interpolation}.
The interpolation results verify the continuity in the attribute space and show that our model can generalize in the distribution, rather than memorize trivial visual information.

\vspace{\paramargin}
\Paragraph{Attribute transfer.} 
We demonstrate the results of the attribute transfer in \figref{cross}.
Thanks to the representation disentanglement of content and attribute, we are able to perform attribute transfer from images of desired attributes, as illustrated in \figref{architecture}(c).
Moreover, since the content space is shared between two domains, we can generate images conditioned on content features encoded from either domain.
Thus our model can achieve not only inter-domain but also intra-domain attribute transfer.
Note that intra-domain attribute transfer is not explicitly involved in the training process.
%

\subsection{Quantitative Evaluation}
 \vspace{\subsecmargin}
\Paragraph{Realism \vs diversity.} 
Here we have the quantitative evaluation on the realism and diversity of the generated images.
We conduct the experiment using winter $\rightarrow$ summer translation with the Yosemite dataset.
For realism, we conduct a user study using pairwise comparison.
Given a pair of images sampled from real images and translated images generated from various methods, users need to answer the question  ``Which image is more realistic?''
For diversity, similar to~\cite{zhu2017bicyclegan}, we use the LPIPS metric~\cite{zhang2018perceptual} to measure the similarity among images.
We compute the distance between 1000 pairs of randomly sampled images translated from 100 real images.

\figref{realism} and \tabref{diversity} show the results of realism and diversity, respectively.
UNIT obtains low realism score, suggesting that their assumption might not be generally applicable.
CycleGAN achieves the highest scores in realism, yet the diversity is limited.
The diversity and the visual quality of Cycle/Bicycle are constrained by the data CycleGAN can generate.
Our results also demonstrate the need for the content discriminator.
%


\vspace{\paramargin}
\Paragraph{Reconstruction ability.} 
In addition to diversity evaluation, we conduct an experiment on the edge-to-shoes dataset to measure the quality of the disentangled encoding.
Our model was trained using unpaired data.
At test time, given a paired data $\{x,y\}$, we can evaluate the quality of content-attribute disentanglement by measuring the reconstruction errors of $y$ with $\hat{y} = G_\mathcal{Y}(E_\mathcal{X}^c(x), E_\mathcal{Y}^a(y))$.

We compare our model with BicycleGAN, which requires paired data during training.
\tabref{recon} shows our model performs comparably with BicycleGAN despite training without paired data.
Moreover, the result suggests that the content discriminator contributes greatly to the quality of disentangled representation.
\vspace{-1mm}
\subsection{Domain Adaptation}
\begin{figure*}[t]
	\centering
    \subfloat[Examples from  MNIST/MNIST-M ]{
        \mpage{0.45}{
		     \mpage{0.45}{MNIST}\hfill\mpage{0.45}{MNIST-M}
             \mpage{0.45}{(Source)}\hfill\mpage{0.45}{(Target)}
             \includegraphics[width=\linewidth]{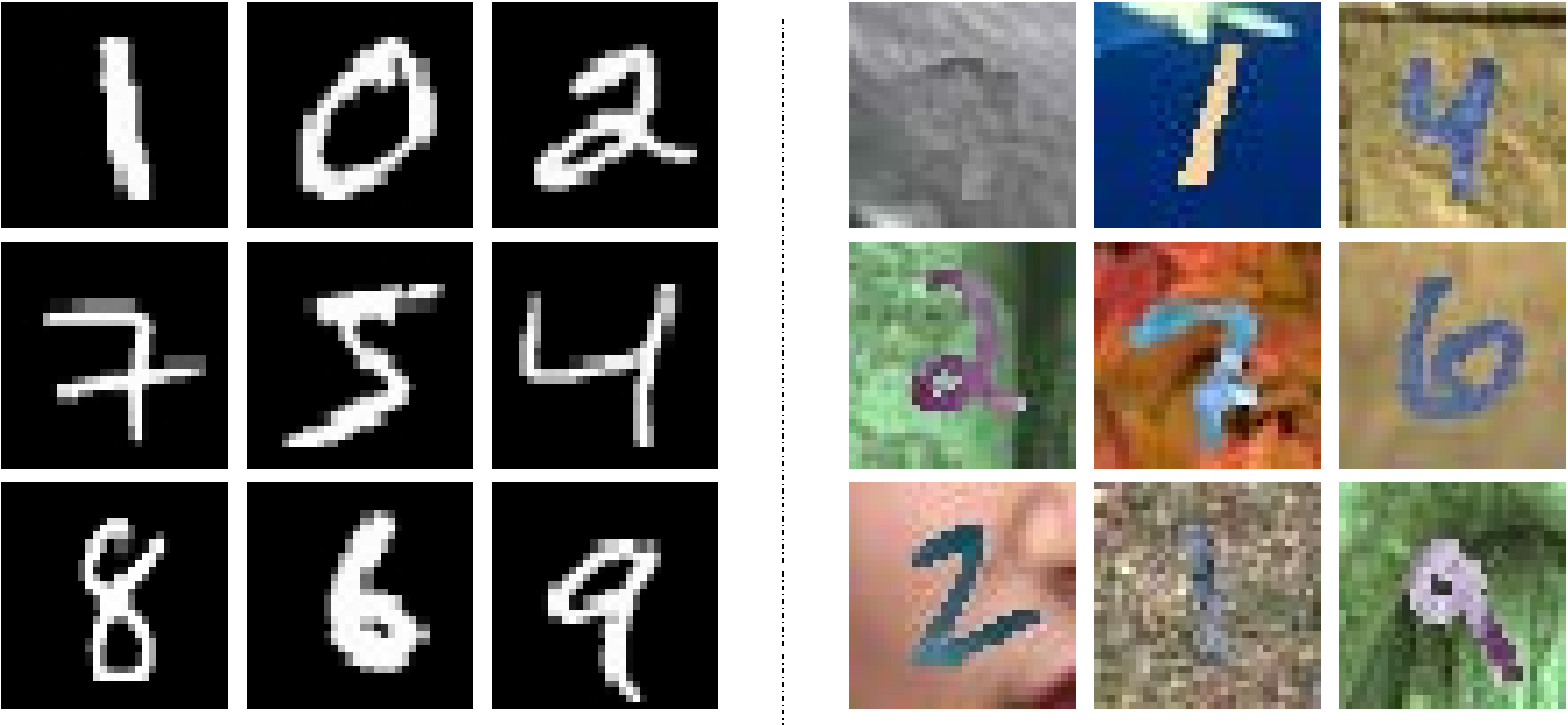}
              \vspace{-4mm}
          }
    }
    \hfill
    \subfloat[Examples from Cropped Linemod ]{
        \mpage{0.45}{
		     \mpage{0.45}{Synthetic}\hfill\mpage{0.45}{Real}
             \mpage{0.45}{(Source)}\hfill\mpage{0.45}{(Target)}
             \includegraphics[width=\linewidth]{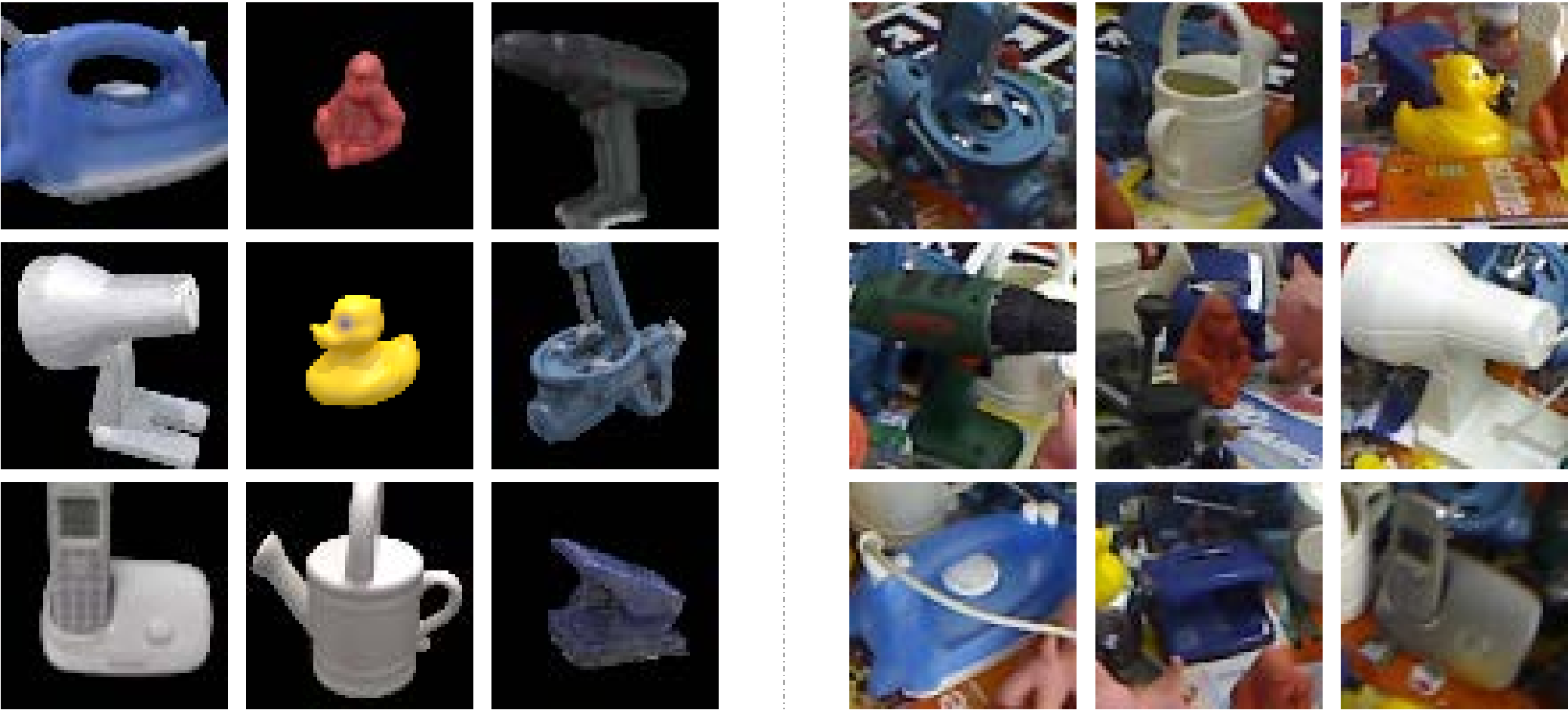}
             \vspace{-4mm}
         }
    }
    \vspace{\subfigmargin}
    
	\subfloat[MNIST $\rightarrow$ MNIST-M]{
        \mpage{0.45}{
		   \mpage{0.15}{Source}\hfill\mpage{0.8}{Generated}
           \includegraphics[width=\linewidth]{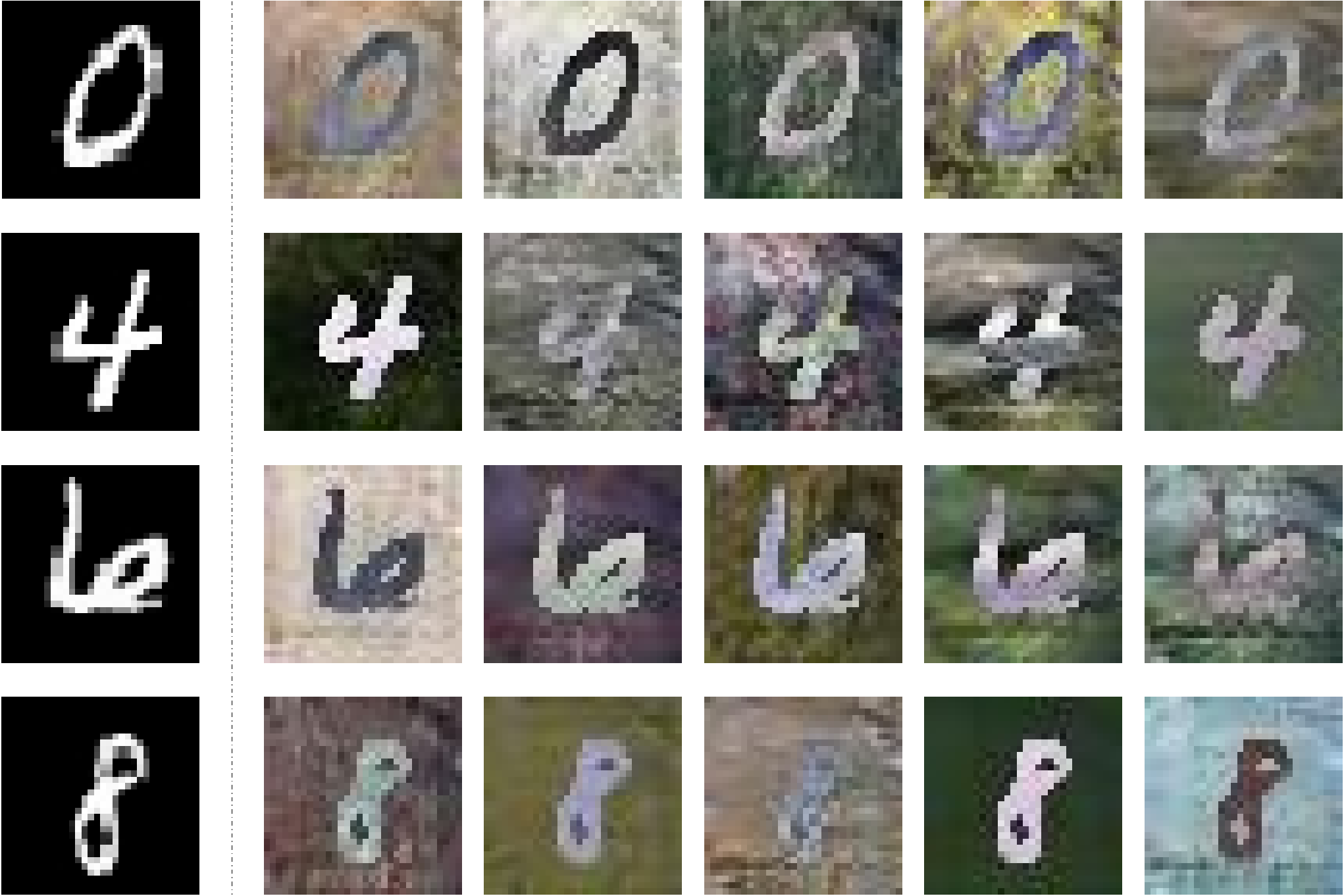}
           \vspace{-4mm}
          }
    }
    \hfill
    \subfloat[Synthetic $\rightarrow$ Real Cropped LineMod]{
        \mpage{0.45}{
		    \mpage{0.15}{Source}\hfill\mpage{0.8}{Generated}
            \includegraphics[width=\linewidth]{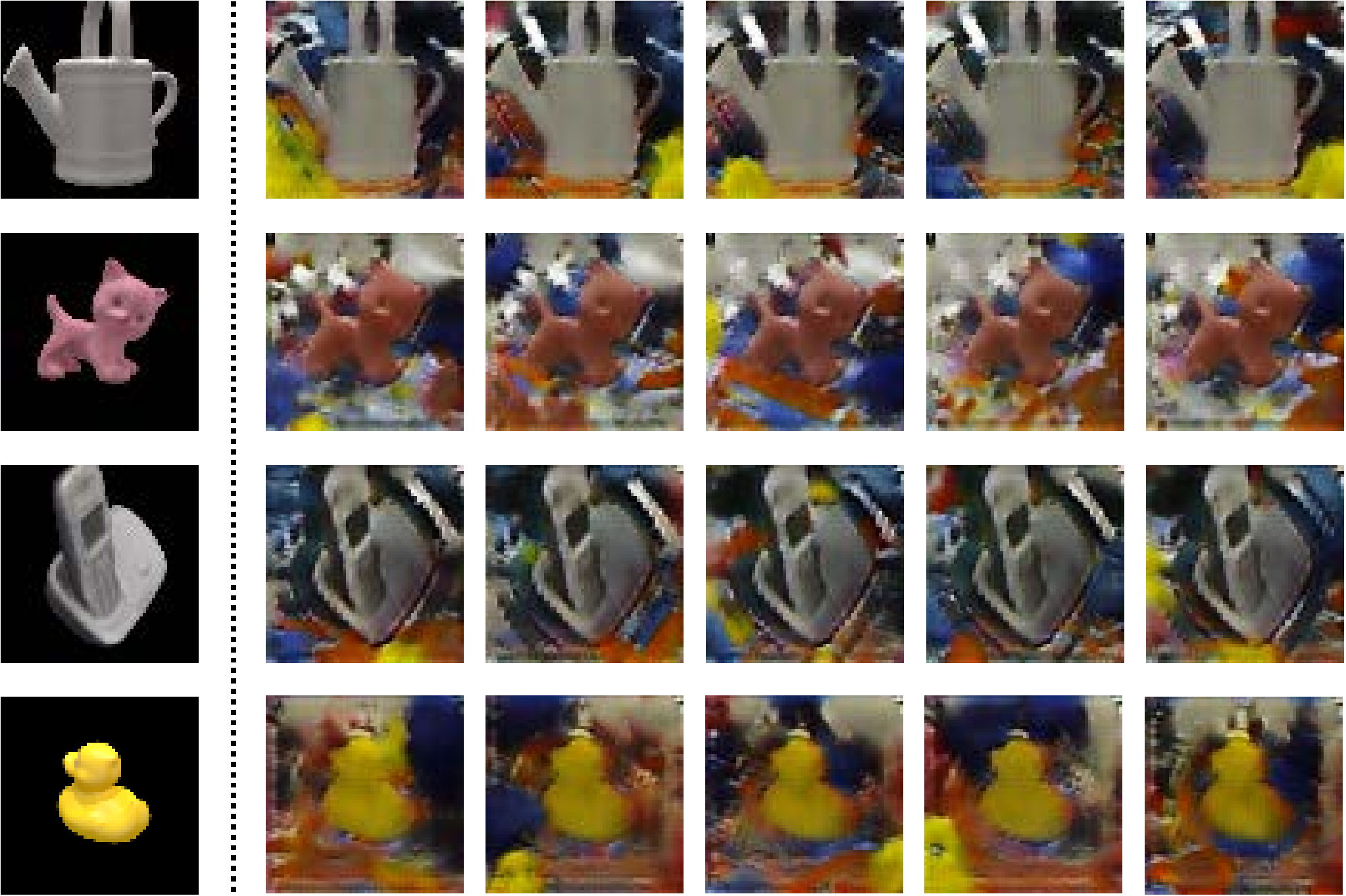}
            \vspace{-4mm}
        }
    }
    \vspace{-1mm}
	\caption{\textbf{Domain adaptation experiments.} We conduct the experiment on (a) MNIST to MNIST-M, and (b) Synthetic to Realistic Cropped LineMod. (c)(d) Our method can generate diverse images that benefit the domain adaptation.
   } 
	\label{figure:da}
    \vspace{\figmargin}
    \vspace{+2mm}
\end{figure*}

\vspace{-1mm}
We demonstrate that the proposed image-to-image translation scheme can benefit unsupervised domain adaptation.
%
%
Following PixelDA~\cite{bousmalis2017unsupervisedda}, we conduct experiments on the classification and pose estimation tasks using MNIST~\cite{lecun1998MNIST} to MNIST-M~\cite{ganin2016MNISTM}, and Synthetic Cropped LineMod to Cropped LineMod~\cite{hinterstoisser2012linemod,wohlhart2015croplinemod}.
Several example images in these datasets are shown in \figref{da} (a) and (b).
To evaluate our method, we first translate the labeled source images to the target domain.
We then treat the generated labeled images as training data and train the classifiers of each task in the target domain.
For a fair comparison, we use the classifiers with the same architecture as PixelDA.
We compare the proposed method with CycleGAN, which generates the most realistic images in the target domain according to our previous experiment, and three state-of-the-art domain adaptation algorithms: PixelDA, DANN~\cite{ganin2016domain} and DSN~\cite{bousmalis2016domain}.
\begin{table}[t]
	\caption{\textbf{Domain adaptation results}. 
    We report the classification accuracy and the pose estimation error on MNIST to MNIST-M and Synthetic Cropped LineMod to Cropped LineMod. 
    The entries ``Source-only" and ``Target-only" represent that the training uses either image only from the source and target domain. 
    Numbers in parenthesis are reported by PixelDA, which are slightly different from what we obtain.
    \jiabin{Maybe use a footnote to talk about this rather in the table caption.}
    }
	\label{tab:da}
	\centering
	\small
    \subfloat[MNIST-M]{
    \vspace{-2mm}
	\begin{tabular}{l cc} 
    	\toprule
		Model & \makecell{Classification \\ Accuracy ($\%$)}\\
        \midrule
        Source-only & 56.6 \\
        \midrule
        CycleGAN~\cite{zhu2017cyclegan} & 74.5 \\
        Ours, $\times 1$ & 86.93\\
        Ours, $\times 3$ & \underline{90.21}\\
        Ours, $\times 5$ & \textbf{91.54}\\
        \midrule
        DANN~\cite{ganin2016domain} & 77.4 \\
        DSN~\cite{bousmalis2016domain} & \underline{83.2} \\
        PixelDA~\cite{bousmalis2017unsupervisedda} & \textbf{95.9} \\
        \midrule
        Target-only & 96.5 \\
        \bottomrule
	\end{tabular}}
    \vspace{-4mm}
    \qquad
    \subfloat[Cropped LineMod]{
    \vspace{-2mm}
	\begin{tabular}{l cc} 
    	\toprule
		Model & \makecell{Classification \\ Accuracy ($\%$)} & \makecell{Mean Angle \\ Error ($\degree$)}\\
        \midrule
        Source-only & 42.9 (47.33) & 73.7 (89.2) \\
        \midrule
        CycleGAN~\cite{zhu2017cyclegan} & 68.18 & 47.45 \\
        Ours, $\times 1$ & 95.91 & 42.06\\
        Ours, $\times 3$ & \underline{97.04} & \underline{37.35}\\
        Ours, $\times 5$ & \textbf{98.12} & \textbf{34.4}\\
        \midrule
        DANN~\cite{ganin2016domain} & \underline{99.9} & 56.58 \\
        DSN~\cite{bousmalis2016domain} & \textbf{100} & \underline{53.27} \\
        PixelDA~\cite{bousmalis2017unsupervisedda} & 99.98 &\textbf{23.5} \\ 
        \midrule
        Target-only & 100 &12.3 (6.47) \\
        \bottomrule
	\end{tabular}}
    \vspace{-4mm}
\end{table}

We present the quantitative comparisons in \tabref{da} and visual results from our method in Figure~\ref{figure:da}(c)(d).
Since our model can generate diverse output, we generate one time, three times, and five times (denoted as $\times 1, \times3, \times 5$) of target images using the same amount of source images.
Our results validate that the proposed method can simulate diverse images in the target domain and improve the performance in target tasks.
While our method does not outperform PixelDA, we note that unlike PixelDA, we do not leverage label information during training.
Compared to CycleGAN, our method performs favorably even with the same amount of generated images (\ie $\times 1$).
We observe that CycleGAN suffers from the mode collapse problem and generates images with similar appearances, which degrade the performance of the adapted classifiers.

\begin{figure*}[t]
	\centering
	\subfloat[Summer $\rightarrow$ Winter]{
    \includegraphics[width=0.4\linewidth]{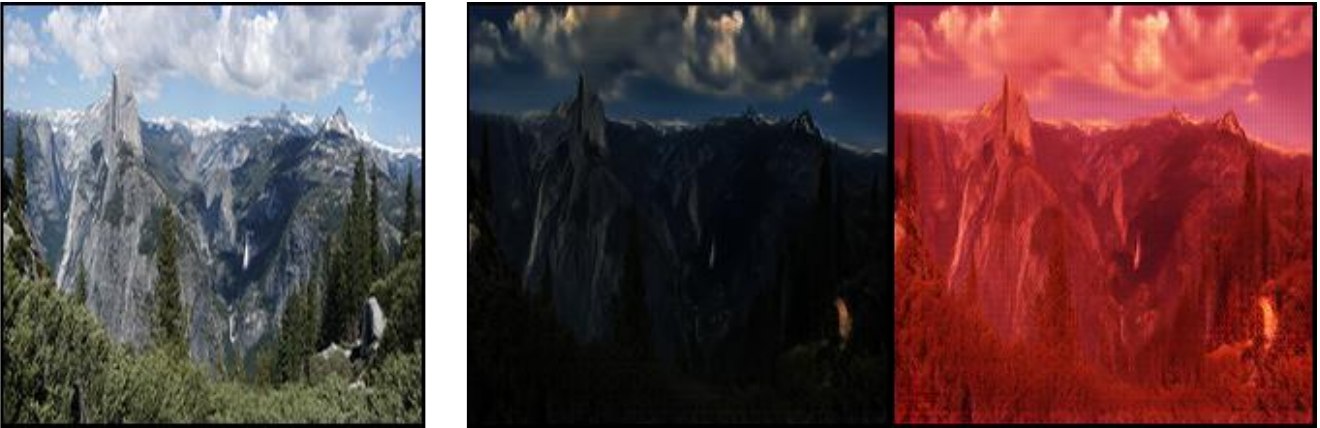}
	}
	\subfloat[van Gogh $\rightarrow$ Monet]{
    \includegraphics[width=0.4\linewidth]{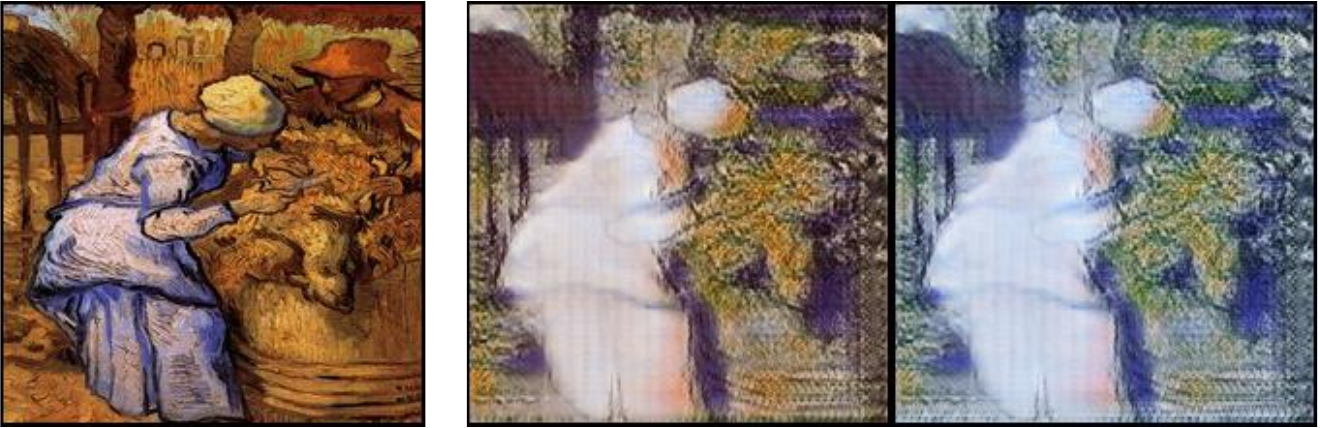}
	}
    \vspace{\subfigmargin}
    \vspace{-3mm}
   
	\caption{\textbf{Failure Cases.} Typical cases: (a) Attribute space not fully exploited. (b) Distribution characteristic difference. }
	\label{figure:failure}
    \vspace{-5mm}
\end{figure*}
\vspace{-1mm}
\vspace{\secmargin}
\subsection{Limitations}
\label{subsec:limitation}
\vspace{\secmargin}
%
Our method has the following limitations.
First, due to the limited amount of training data, the attribute space is not fully exploited.
Our I2I translation fails when the sampled attribute vectors locate in under-sampled space, see \figref{failure}(a).
Second, it remains difficult when the domain characteristics differ significantly.
For example, \figref{failure}(b) shows a failure case on the human figure due to the lack of human-related portraits in Monet collections.
\vspace{-1mm}
\section{Conclusions}
\label{sec:conclusion}
\vspace{\secmargin}
In this paper, we present a novel disentangled representation framework for diverse image-to-image translation with unpaired data.
we propose to disentangle the latent space to a content space that encodes common information between domains, and a domain-specific attribute space that can model the diverse variations given the same content.
We apply a content discriminator to facilitate the representation disentanglement.
We propose a cross-cycle consistency loss for cyclic reconstruction to train in the absence of paired data.
Qualitative and quantitative results show that the proposed model produces realistic and diverse images.
We also apply the proposed method to domain adaptation and achieve competitive performance compared to the state-of-the-art methods.
\vspace{-1mm}
\section*{Acknowledgements}
\vspace{-3mm}
This work is supported in part by the NSF CAREER Grant \#1149783, the NSF Grant \#1755785, and gifts from Verisk, Adobe and Nvidia.

\clearpage

\bibliographystyle{splncs04}
\bibliography{eccv18}
\end{document}

%% file: macro.tex
\usepackage{xspace}
\usepackage{color}
\usepackage{multirow}
\usepackage{ifthen}

\graphicspath{{figures/}}

\newcommand{\tb}[1]{\textbf{#1}}

\newcommand {\minghsuan}[1]{{\color{red}\bf{Ming-Hsuan: }#1}\normalfont}
\newcommand {\james}[1]{{\color{magenta}\textbf{James: }#1}\normalfont}
\newcommand {\jiabin}[1]{{\color{blue}\textbf{Jia-Bin: }#1}\normalfont}

\long\def\ignorethis#1{}

\definecolor{gray}{rgb}{0.35,0.35,0.35}
\definecolor{MyBlue}{rgb}{0,0.2,0.8}
\definecolor{MyRed}{rgb}{0.8,0.2,0}
\definecolor{MyGreen}{rgb}{0.0,0.5,0.1}
\definecolor{MyGray}{rgb}{0.4,0.4,0.4}

\newlength\paramargin
\newlength\figmargin
\newlength\subfigmargin
\newlength\secmargin
\newlength\subsecmargin
\newlength\tabmargin
\newlength\eqmargin

\usepackage{array}
\newcolumntype{L}[1]{>{\raggedright\let\newline\\\arraybackslash\hspace{0pt}}m{#1}}
\newcolumntype{C}[1]{>{\centering\let\newline\\\arraybackslash\hspace{0pt}}m{#1}}
\newcolumntype{R}[1]{>{\raggedleft\let\newline\\\arraybackslash\hspace{0pt}}m{#1}}

\newcommand{\mpage}[2]
{
\begin{minipage}[t]{#1\linewidth}\centering
#2
\end{minipage}
}

\newcommand{\Paragraph}[1]
{\vspace{2mm} \noindent \textbf{#1}}

\def\ie{i.e.,~}
\def\eg{e.g.,~}

\def\vs{vs.~}
\def\etal{et~al.\xspace}

\newcommand{\subsecref}[1]{Section~\ref{subsec:#1}}
\newcommand{\figref}[1]{Figure~\ref{figure:#1}}
\newcommand{\tabref}[1]{Table~\ref{tab:#1}}


%% file: eccv18_multimodal_arxiv.bbl
\begin{thebibliography}{10}
\providecommand{\url}[1]{\texttt{#1}}
\providecommand{\urlprefix}{URL }
\providecommand{\doi}[1]{https://doi.org/#1}

\bibitem{almahairi2018augmented}
Almahairi, A., Rajeswar, S., Sordoni, A., Bachman, P., Courville, A.: Augmented
  cyclegan: Learning many-to-many mappings from unpaired data. arXiv preprint
  arXiv:1802.10151  (2018)

\bibitem{arjovsky2017wgan}
Arjovsky, M., Chintala, S., Bottou, L.: Wasserstein {GAN}. In: ICML (2017)

\bibitem{bousmalis2017unsupervisedda}
Bousmalis, K., Silberman, N., Dohan, D., Erhan, D., Krishnan, D.: Unsupervised
  pixel-level domain adaptation with generative adversarial networks. In: CVPR
  (2017)

\bibitem{bousmalis2016domain}
Bousmalis, K., Trigeorgis, G., Silberman, N., Krishnan, D., Erhan, D.: Domain
  separation networks. In: NIPS (2016)

\bibitem{cao2018dida}
Cao, J., Katzir, O., Jiang, P., Lischinski, D., Cohen-Or, D., Tu, C., Li, Y.:
  Dida: Disentangled synthesis for domain adaptation. arXiv preprint
  arXiv:1805.08019  (2018)

\bibitem{chen2017photographic}
Chen, Q., Koltun, V.: Photographic image synthesis with cascaded refinement
  networks. In: ICCV (2017)

\bibitem{chen2016infogan}
Chen, X., Duan, Y., Houthooft, R., Schulman, J., Sutskever, I., Abbeel, P.:
  Info{GAN}: Interpretable representation learning by information maximizing
  generative adversarial nets. In: NIPS (2016)

\bibitem{cheung2014discovering}
Cheung, B., Livezey, J.A., Bansal, A.K., Olshausen, B.A.: Discovering hidden
  factors of variation in deep networks. In: ICLR workshop (2015)

\bibitem{choi2017stargan}
Choi, Y., Choi, M., Kim, M., Ha, J.W., Kim, S., Choo, J.: Stargan: Unified
  generative adversarial networks for multi-domain image-to-image translation.
  In: CVPR. vol.~1711 (2018)

\bibitem{denton2017unsupervised}
Denton, E.L., Birodkar, V.: Unsupervised learning of disentangled
  representations from video. In: NIPS (2017)

\bibitem{ganin2015unsupervised}
Ganin, Y., Lempitsky, V.: Unsupervised domain adaptation by backpropagation.
  In: ICML (2015)

\bibitem{ganin2016MNISTM}
Ganin, Y., Ustinova, E., Ajakan, H., Germain, P., Larochelle, H., Laviolette,
  F., Marchand, M., Lempitsky, V.: Domain-adversarial training of neural
  networks. JMLR  (2016)

\bibitem{ganin2016domain}
Ganin, Y., Ustinova, E., Ajakan, H., Germain, P., Larochelle, H., Laviolette,
  F., Marchand, M., Lempitsky, V.: Domain-adversarial training of neural
  networks. JMLR  (2016)

\bibitem{goodfellow2014GAN}
Goodfellow, I., Pouget-Abadie, J., Mirza, M., Xu, B., Warde-Farley, D., Ozair,
  S., Courville, A., Bengio, Y.: Generative adversarial nets. In: NIPS (2014)

\bibitem{hinterstoisser2012linemod}
Hinterstoisser, S., Lepetit, V., Ilic, S., Holzer, S., Bradski, G., Konolige,
  K., Navab, N.: Model based training, detection and pose estimation of
  texture-less 3d objects in heavily cluttered scenes. In: ACCV (2012)

\bibitem{hoffman2017cycada}
Hoffman, J., Tzeng, E., Park, T., Zhu, J.Y., Isola, P., Saenko, K., Efros,
  A.A., Darrell, T.: {CyCADA}: Cycle-consistent adversarial domain adaptation.
  In: ICML (2018)

\bibitem{huang2018munit}
Huang, X., Liu, M.Y., Belongie, S., Kautz, J.: Multimodal unsupervised
  image-to-image translation. In: ECCV (2018)

\bibitem{isola2017pix2pix}
Isola, P., Zhu, J.Y., Zhou, T., Efros, A.A.: Image-to-image translation with
  conditional adversarial networks. In: CVPR (2017)

\bibitem{kim2017discogan}
Kim, T., Cha, M., Kim, H., Lee, J., Kim, J.: Learning to discover cross-domain
  relations with generative adversarial networks. In: ICML (2017)

\bibitem{kinga2015adam}
Kinga, D., Adam, J.B.: A method for stochastic optimization. In: ICLR (2015)

\bibitem{kingma2014semi}
Kingma, D.P., Rezende, D., Mohamed, S.J., Welling, M.: Semi-supervised learning
  with deep generative models. In: NIPS (2014)

\bibitem{lai2017deep}
Lai, W.S., Huang, J.B., Ahuja, N., Yang, M.H.: Deep laplacian pyramid networks
  for fast and accurate superresolution. In: CVPR (2017)

\bibitem{larsson2016colorization}
Larsson, G., Maire, M., Shakhnarovich, G.: Learning representations for
  automatic colorization. In: ECCV (2016)

\bibitem{lecun1998MNIST}
LeCun, Y., Bottou, L., Bengio, Y., Haffner, P.: Gradient-based learning applied
  to document recognition. Proceedings of the IEEE  (1998)

\bibitem{ledig2016photo}
Ledig, C., Theis, L., Husz{\'a}r, F., Caballero, J., Cunningham, A., Acosta,
  A., Aitken, A., Tejani, A., Totz, J., Wang, Z., Shi, W.: Photo-realistic
  single image super-resolution using a generative adversarial network. In:
  CVPR (2017)

\bibitem{li2016deep}
Li, Y., Huang, J.B., Ahuja, N., Yang, M.H.: Deep joint image filtering. In:
  ECCV (2016)

\bibitem{liu2017unit}
Liu, M.Y., Breuel, T., Kautz, J.: Unsupervised image-to-image translation
  networks. In: NIPS (2017)

\bibitem{liu2015celeb}
Liu, Z., Luo, P., Wang, X., Tang, X.: Deep learning face attributes in the
  wild. In: ICCV (2015)

\bibitem{ma2018exemplar}
Ma, L., Jia, X., Georgoulis, S., Tuytelaars, T., Van~Gool, L.: Exemplar guided
  unsupervised image-to-image translation. arXiv preprint arXiv:1805.11145
  (2018)

\bibitem{makhzani2015adversarial}
Makhzani, A., Shlens, J., Jaitly, N., Goodfellow, I., Frey, B.: Adversarial
  autoencoders. In: ICLR workshop (2016)

\bibitem{mathieu2016disentangling}
Mathieu, M., Zhao, J., Sprechmann, P., Ramesh, A., LeCun, Y.: Disentangling
  factors of variation in deep representation using adversarial training. In:
  NIPS (2016)

\bibitem{murez2018image}
Murez, Z., Kolouri, S., Kriegman, D., Ramamoorthi, R., Kim, K.: Image to image
  translation for domain adaptation. In: CVPR (2018)

\bibitem{paszke2017pytorch}
Paszke, A., Gross, S., Chintala, S., Chanan, G., Yang, E., DeVito, Z., Lin, Z.,
  Desmaison, A., Antiga, L., Lerer, A.: Automatic differentiation in pytorch.
  In: NIPS workshop (2017)

\bibitem{radford2016dcgan}
Radford, A., Metz, L., Chintala, S.: Unsupervised representation learning with
  deep convolutional generative adversarial networks. In: ICLR (2016)

\bibitem{reed2016text2img}
Reed, S., Akata, Z., Yan, X., Logeswaran, L., Schiele, B., Lee, H.: Generative
  adversarial text to image synthesis. In: ICML (2016)

\bibitem{shrivastava2017apple}
Shrivastava, A., Pfister, T., Tuzel, O., Susskind, J., Wang, W., Webb, R.:
  Learning from simulated and unsupervised images through adversarial training.
  In: CVPR (2017)

\bibitem{sun2016return}
Sun, B., Feng, J., Saenko, K.: Return of frustratingly easy domain adaptation.
  In: AAAI (2016)

\bibitem{taigman2016unsupervised}
Taigman, Y., Polyak, A., Wolf, L.: Unsupervised cross-domain image generation.
  In: ICLR (2017)

\bibitem{Tsai_adaptseg_2018}
Tsai, Y.H., Hung, W.C., Schulter, S., Sohn, K., Yang, M.H., Chandraker, M.:
  Learning to adapt structured output space for semantic segmentation. In: CVPR
  (2018)

\bibitem{tzeng2014deep}
Tzeng, E., Hoffman, J., Zhang, N., Saenko, K., Darrell, T.: Deep domain
  confusion: Maximizing for domain invariance. arXiv preprint arXiv:1412.3474
  (2014)

\bibitem{vondrick2016videogan}
Vondrick, C., Pirsiavash, H., Torralba, A.: Generating videos with scene
  dynamics. In: NIPS (2016)

\bibitem{wang2017pix2pixhd}
Wang, T.C., Liu, M.Y., Zhu, J.Y., Tao, A., Kautz, J., Catanzaro, B.:
  High-resolution image synthesis and semantic manipulation with conditional
  gans. In: CVPR (2018)

\bibitem{wohlhart2015croplinemod}
Wohlhart, P., Lepetit, V.: Learning descriptors for object recognition and 3d
  pose estimation. In: CVPR (2015)

\bibitem{yi2017dualgan}
Yi, Z., Zhang, H.R., Tan, P., Gong, M.: Dualgan: Unsupervised dual learning for
  image-to-image translation. In: ICCV (2017)

\bibitem{Yu2014edge2shoe}
Yu, A., Grauman, K.: Fine-grained visual comparisons with local learning. In:
  CVPR (2014)

\bibitem{zhang2016colorization2}
Zhang, R., Isola, P., Efros, A.A.: Colorful image colorization. In: ECCV (2016)

\bibitem{zhang2018perceptual}
Zhang, R., Isola, P., Efros, A.A., Shechtman, E., Wang, O.: The unreasonable
  effectiveness of deep networks as a perceptual metric. In: CVPR (2018)

\bibitem{zhu2017cyclegan}
Zhu, J.Y., Park, T., Isola, P., Efros, A.A.: Unpaired image-to-image
  translation using cycle-consistent adversarial networks. In: ICCV (2017)

\bibitem{zhu2017bicyclegan}
Zhu, J.Y., Zhang, R., Pathak, D., Darrell, T., Efros, A.A., Wang, O.,
  Shechtman, E.: Toward multimodal image-to-image translation. In: NIPS (2017)

\end{thebibliography}
